%% file: root.tex
\documentclass[lettersize,journal]{IEEEtran}
\usepackage{amsmath,amsfonts}
\usepackage{algorithmic}
\usepackage{algorithm}
\usepackage{array}
\usepackage[caption=false,font=normalsize,labelfont=sf,textfont=sf]{subfig}
\usepackage{textcomp}
\usepackage{stfloats}
\usepackage{url}
\usepackage{verbatim}
\usepackage{graphicx}
\usepackage{cite}
\usepackage{amssymb}  
\usepackage{adjustbox}
\usepackage{gensymb}
\usepackage{threeparttable}
\usepackage{multirow}
\usepackage{booktabs}
\usepackage[table]{xcolor}
\begin{document}

\title{\textbf{\texttt{RAVE}}: End-to-end Hierarchical Visual Localization with \\
Rasterized and Vectorized HD map}

\author{Jinyu Miao$^{1}$, Tuopu Wen$^{1,*}$, Kun Jiang$^{1,*}$, Kangan Qian$^{1}$, Zheng Fu$^{1}$, Yunlong Wang$^{2}$, \\
Zhihuang Zhang$^{3}$, Mengmeng Yang$^{1}$, Jin Huang$^{1}$, Zhihua Zhong$^{4}$, Diange Yang$^{1,*}$ 
\thanks{This work was supported in part by the National Natural Science Foundation of China (52394264, 52472449, U22A20104, 52372414, 52402499), Beijing Natural Science Foundation (23L10038, L231008), Beijing Municipal Science and Technology Commission (Z241100003524013, Z241100003524009), and China Postdoctoral Science Foundation (2024M761636).}
\thanks{$^{1}$Jinyu Miao, Tuopu Wen, Kun Jiang, Kangan Qian, Zheng Fu, Mengmeng Yang, Jin Huang, and Diange Yang are with the School of Vehicle and Mobility, and State Key Laboratory of Intelligent Green Vehicle and Mobility, Tsinghua University, Beijing, China.}%
\thanks{$^{2}$Yunlong Wang is with NIO Inc., Beijing, China}%
\thanks{$^{3}$Zhihuang Zhang is with Qcraft Inc., Beijing, China}%
\thanks{$^{4}$Zhihua Zhong is with Chinese Academy of Engineering, Beijing, China}%
\thanks{$^{*}$Corresponding author: Diange Yang, Kun Jiang, and Tuopu Wen}%
}

\markboth{Journal of \LaTeX\ Class Files,~Vol.~14, No.~8, August~2021}%
{Shell \MakeLowercase{\textit{et al.}}: A Sample Article Using IEEEtran.cls for IEEE Journals}


\maketitle

\input{secs/0-abstract}


\input{secs/1-introduction}

\input{secs/2-relatedwork}
\input{secs/3-method}
\input{secs/4-experiment}
\input{secs/5-conclusion}



\bibliographystyle{IEEEtran}
\bibliography{ref}

\vspace{-33pt}

\begin{IEEEbiography}[{\includegraphics[width=1in,height=1.25in,clip,keepaspectratio]{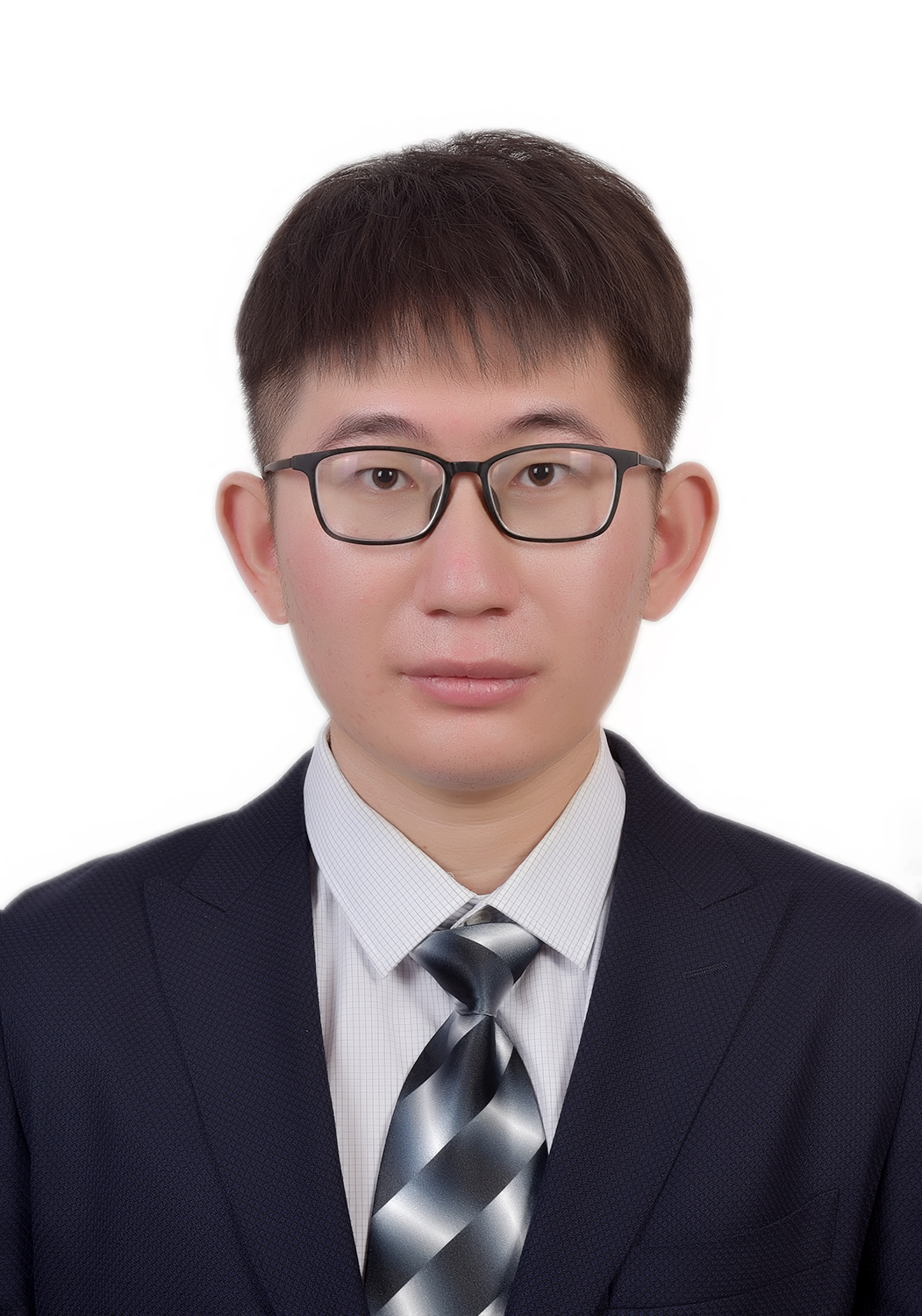}}]{Jinyu Miao}
received his B.S. degree in Automation and his M.S. degree in Control Science and Engineering from the School of Automation Science and Electrical Engineering, Beihang University, Beijing, China, in 2019 and 2022, respectively. He is now currently working toward his Ph.D. degree at the School of Vehicle and Mobility, Tsinghua University, Beijing, China. His research interests include high-precision localization, vehicle dynamics, and planning for autonomous driving.
\end{IEEEbiography}

\vspace{-33pt}

\begin{IEEEbiography}[{\includegraphics[width=1in,height=1.25in,clip,keepaspectratio]{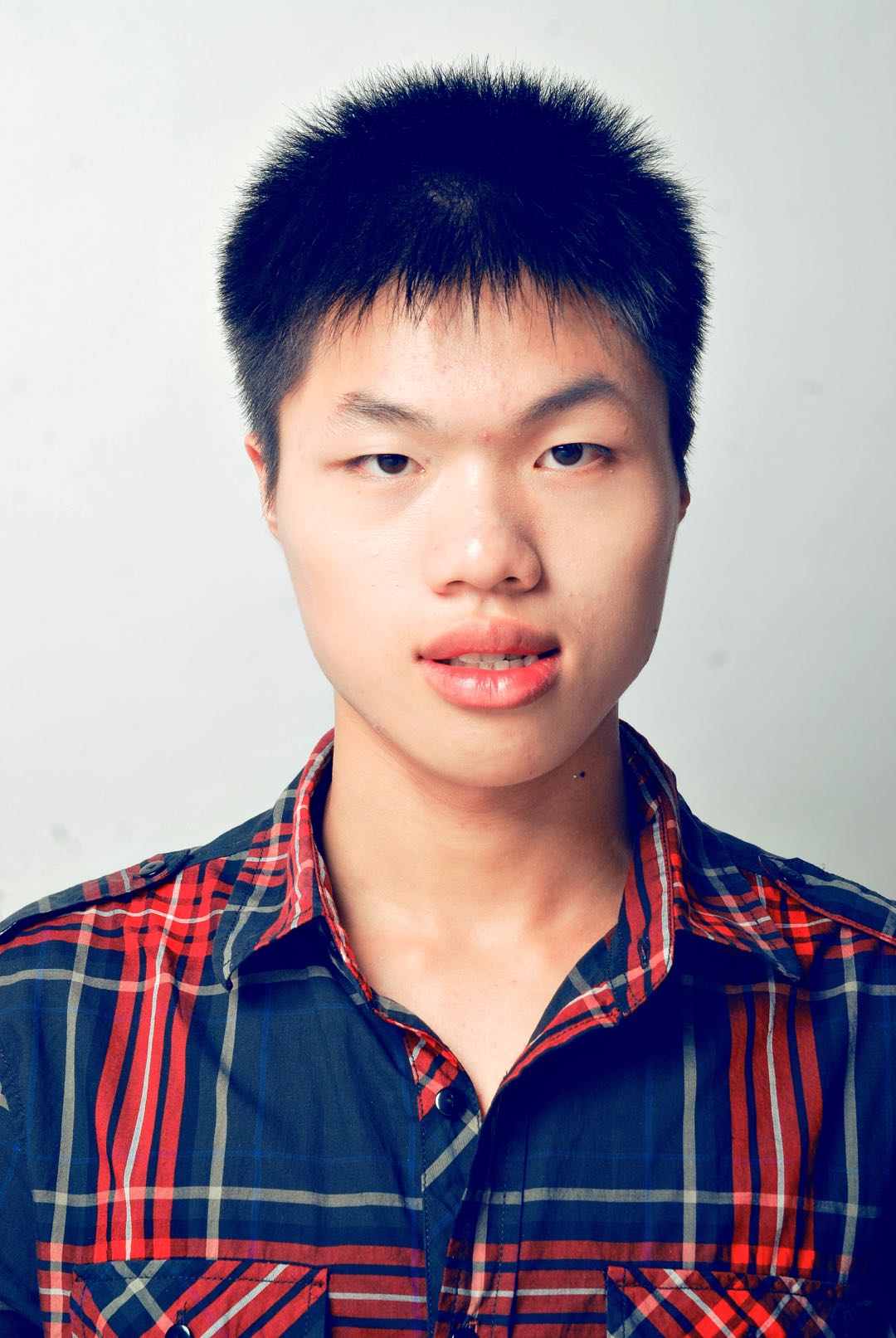}}]{Tuopu Wen}
received the B.S. degree from Electronic Engineering, Tsinghua University, Beijing, China in 2018 and his Ph.D. degree at the School of Vehicle and Mobility of Tsinghua University in 2023. He is now working as a postdoctoral researcher in the School of Vehicle and Mobility of Tsinghua University. His research interests include computer vision, high definition map, and high-precision localization for autonomous driving.
\end{IEEEbiography}

\vspace{-33pt}

\begin{IEEEbiography}[{\includegraphics[width=1in,height=1.25in,clip,keepaspectratio]{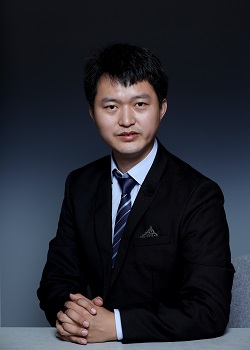}}]{Kun Jiang}
received the B.S. degree in mechanical and automation engineering from Shanghai Jiao Tong University, Shanghai, China in 2011. He received the Master degree in mechatronics system and the Ph.D. degree in information and systems technologies from University of Technology of Compi\`egne (UTC), Compi\`egne, France, in 2013 and 2016, respectively. He is currently an associate research professor at Tsinghua University, Beijing, China. His research interests include autonomous vehicles, high precision map, and sensor fusion.
\end{IEEEbiography}

\vspace{-33pt}

\begin{IEEEbiography}[{\includegraphics[width=1in,height=1.25in,clip,keepaspectratio]{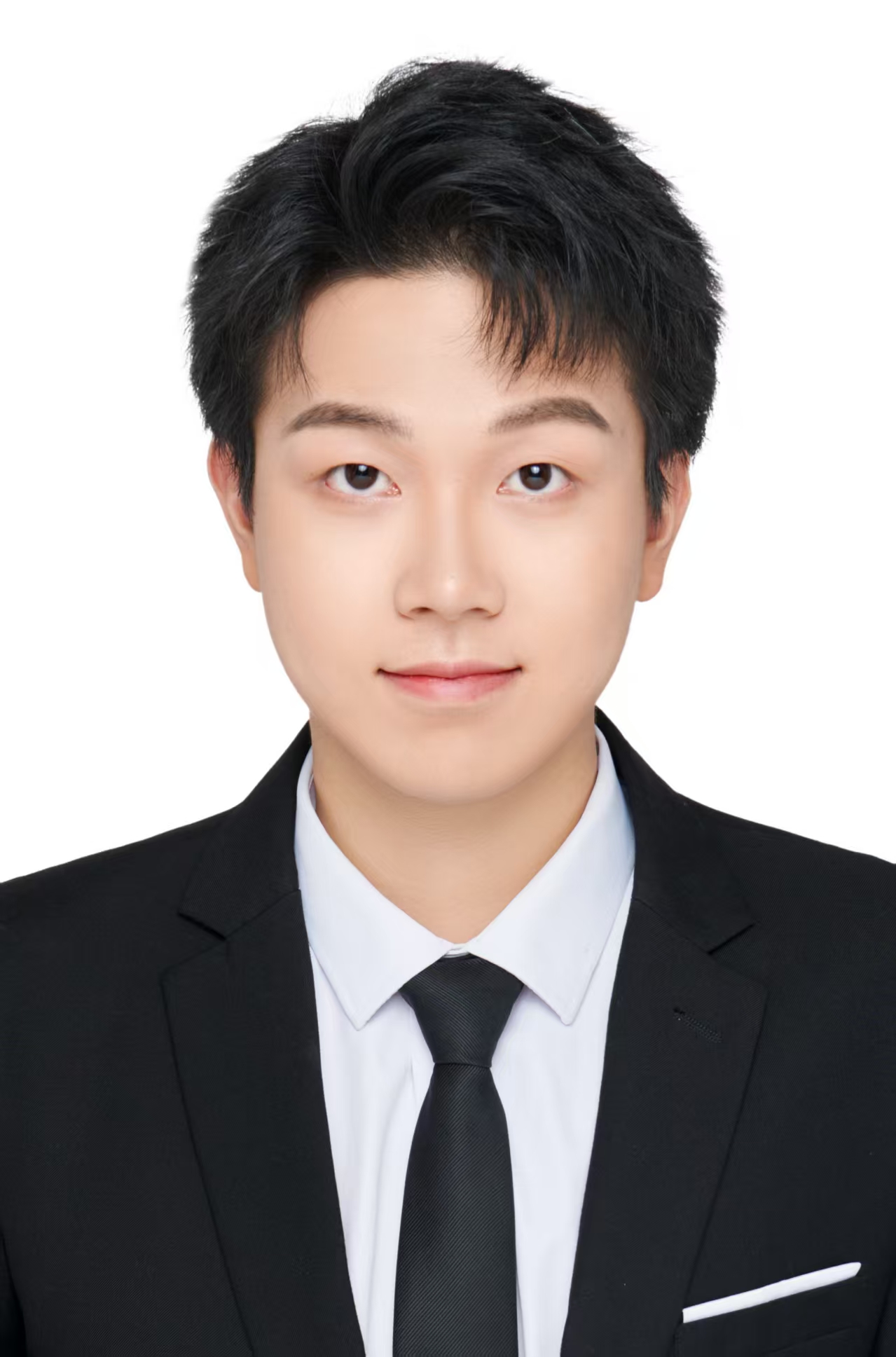}}]{Kangan Qian} 
received his B.S. degree in Mechanical Design, Manufacturing and Automation from Central South University, Changsha, China, in 2023. He is now currently working toward his M.S. degree in Mechanical Engineering at the School of Vehicle and Mobility, Tsinghua University, Beijing, China. His research interests include multimodal perception and reasoning.

\end{IEEEbiography}

\vspace{-33pt}

\begin{IEEEbiography}[{\includegraphics[width=1in,height=1.25in,clip,keepaspectratio]{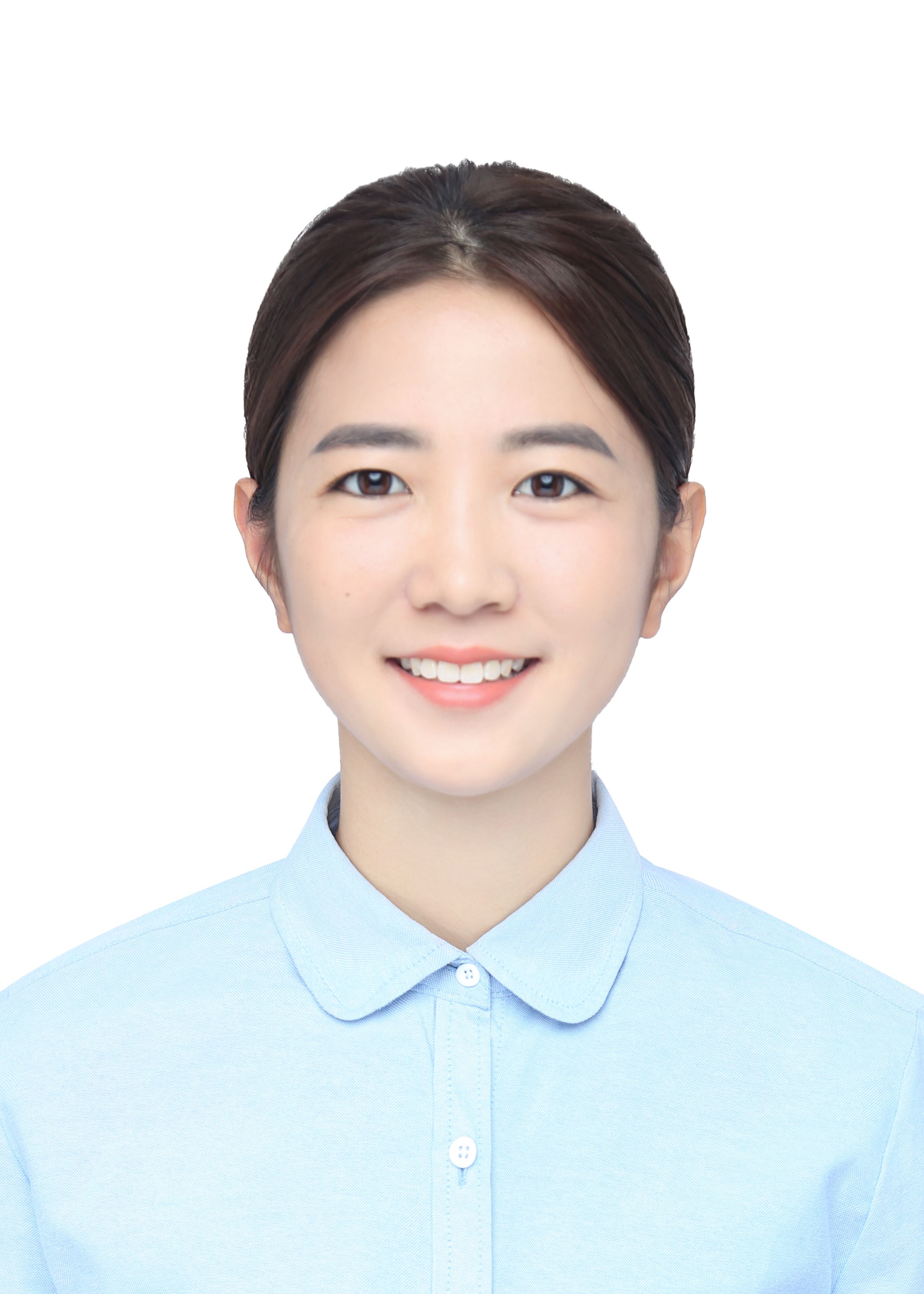}}]{Zheng Fu}
received the M.S. degree in pattern recognition and intelligent systems from the Nanjing University of Posts and Telecommunications, Jiangsu, China, in 2019. She is pursuing the Ph.D. degree at the School of Vehicle and Mobility, Tsinghua University, Beijing, China. Her current research interests include human 3D pose estimation, human intention analysis, and pedestrian trajectory prediction for autonomous driving.
\end{IEEEbiography}

\vspace{-33pt}

\begin{IEEEbiography}[{\includegraphics[width=1in,height=1.25in,clip,keepaspectratio]{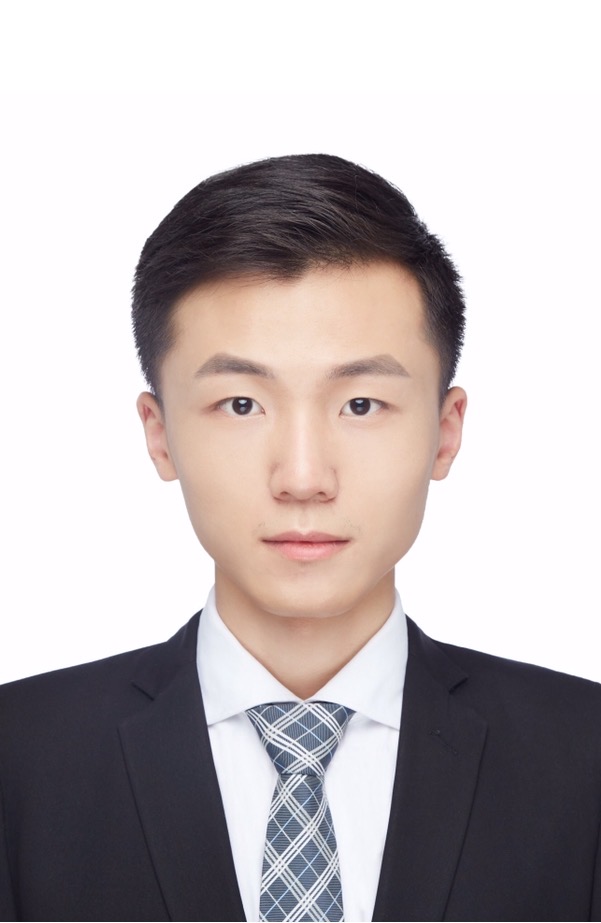}}]{Yunlong Wang}
received the B.S. degree from Electronic Engineering, Tsinghua University, Beijing, China in 2019. He is currently working toward the Ph.D. degree at the School of Vehicle and Mobility of Tsinghua University, Bejing, China. His research interests include LiDAR point cloud semantic segmentation, motion prediction and 3D object detection for autonomous driving.
\end{IEEEbiography}

\vfill

\newpage

\begin{IEEEbiography}[{\includegraphics[width=1in,height=1.25in,clip,keepaspectratio]{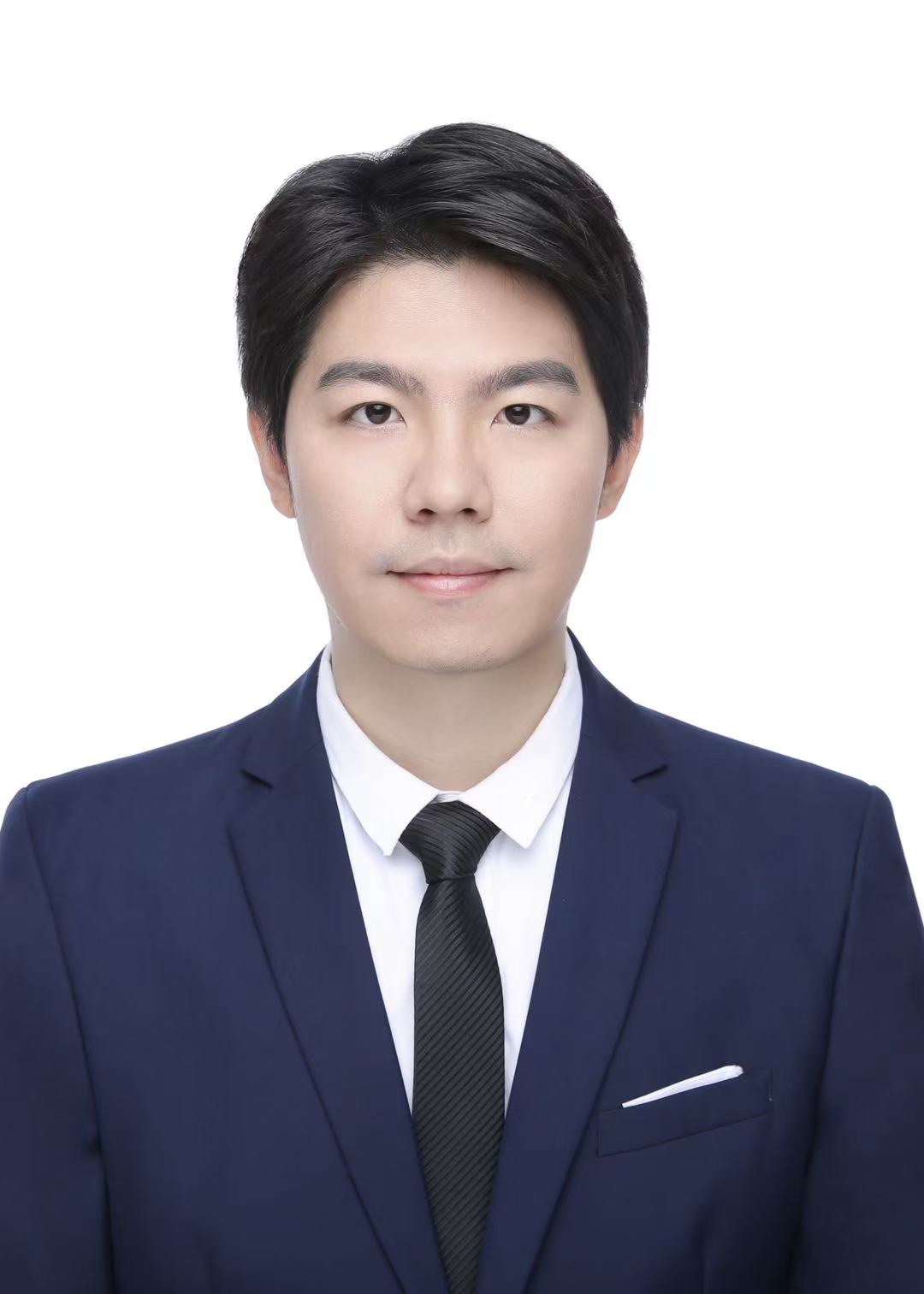}}]{Zhihuang Zhang}
received the B.E. degree in vehicle and mobility from Tsinghua University, Beijing, China, in 2018, and the Ph.D. degree in vehicle and mobility from Tsinghua University, Beijing, China, in 2023. His research interests include learning-based state estimation, time-series analysis, multi-sensor fusion, and high-precision localization for autonomous driving.

\end{IEEEbiography}

\vspace{-33pt}

\begin{IEEEbiography}[{\includegraphics[width=1in,height=1.25in,clip,keepaspectratio]{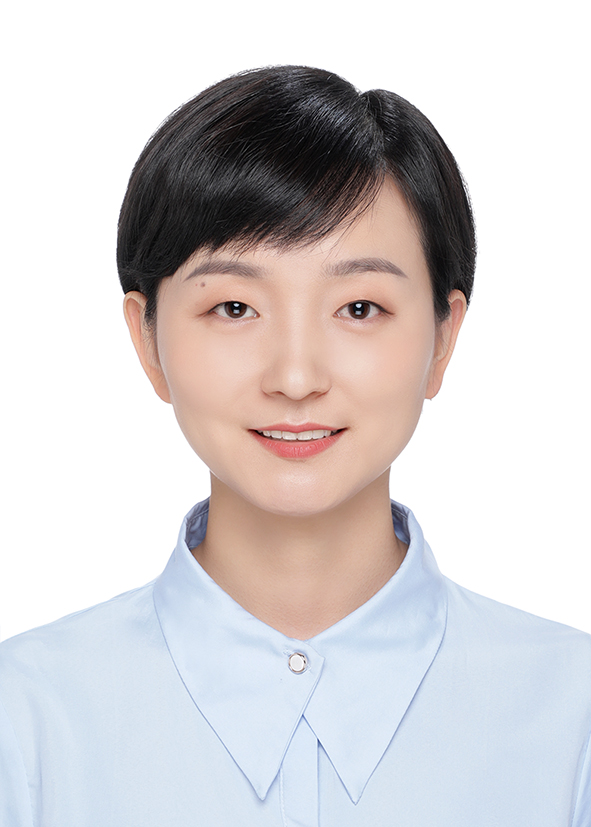}}]{Mengmeng Yang}
received the Ph.D. degree in Photogrammetry and Remote Sensing from Wuhan University, Wuhan, China in 2018. She is currently an assistant research professor at Tsinghua University, Beijing, China. Her research interests include autonomous vehicles, high precision digital map, and sensor fusion. 
\end{IEEEbiography}

\vspace{-33pt}

\begin{IEEEbiography}[{\includegraphics[width=1in,height=1.25in,clip,keepaspectratio]{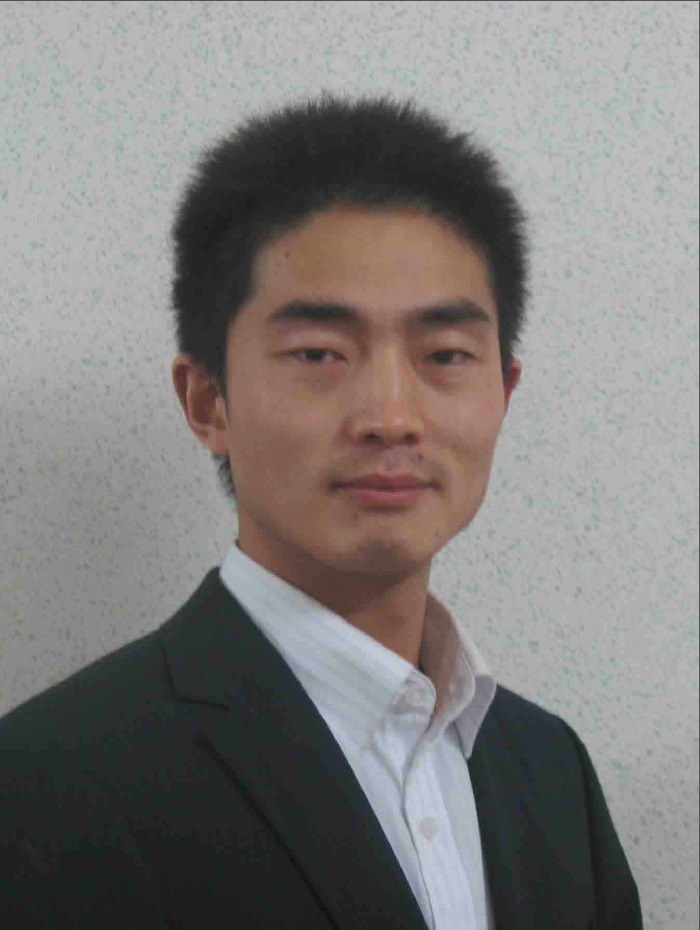}}]{Jin Huang}
 received the B.E. and Ph.D. degrees from the College of Mechanical and Vehicle Engineering, Hunan University, Changsha, China, in 2006 and 2012, respectively. From 2009 to 2011, he was a joint Ph.D. with the George W. Woodruff School of
 Mechanical Engineering, Georgia Institute of Technology, Atlanta, GA, USA. Since 2013 and 2016, he has been the Postdoc and an Assistant Research Professor with Tsinghua University, Beijing, China. His research interests include artificial intelligence in intelligent transportation systems, dynamics control, and fuzzy engineering.
\end{IEEEbiography}

\vspace{-33pt}

\begin{IEEEbiography}[{\includegraphics[width=1in,height=1.25in,clip,keepaspectratio]{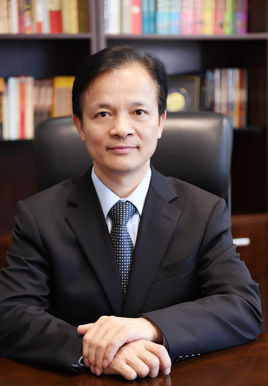}}]{Zhihua Zhong}
received the Ph.D. degree in engineering from Linköping University, Linköping, Sweden, in 1988. He is currently a Professor with the School of Vehicle and Mobility, Tsinghua University, Beijing, China. He was an Elected Member of the Chinese Academy of Engineering in 2005. His research interests include auto collision security technology, the punching and shaping technologies of the auto body, modularity and light-weighing auto technologies, and vehicle dynamics.
\end{IEEEbiography}

\vspace{-33pt}

\begin{IEEEbiography}[{\includegraphics[width=1in,height=1.25in,clip,keepaspectratio]{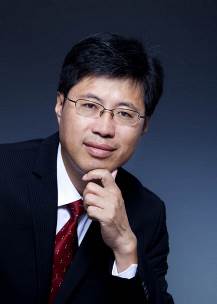}}]{Diange Yang}
received his Ph.D. in Automotive Engineering from Tsinghua University in 2001. He is now a Professor at the school of vehicle and mobility at Tsinghua University. He currently serves as the director of the Tsinghua University Development \& Planning Division. His research interests include autonomous driving, environment perception, and HD map.
He has more than 180 technical publications and 100 patents. He received the National Technology Invention Award and Science and Technology Progress Award in 2010, 2013, 2018, and the Special Prize for Progress in Science and Technology of China Automobile Industry in 2019. 
\end{IEEEbiography}

\vfill

\end{document}

%% file: secs/0-abstract.tex
\begin{abstract}

Accurate localization serves as an important component in autonomous driving systems.
Traditional rule-based localization involves many standalone modules, which is theoretically fragile and requires costly hyperparameter tuning, therefore sacrificing the accuracy and generalization. 
In this paper, we propose an end-to-end visual localization system, \texttt{RAVE}, in which the surrounding images are associated with the HD map data to estimate pose.
To ensure high-quality observations for localization, a low-rank flow-based prior fusion module (\texttt{FLORA}) is developed to incorporate misaligned map prior into the perceived BEV features.
Pursuing a balance among efficiency, interpretability, and accuracy, a hierarchical localization module is proposed, which efficiently estimates poses through a decoupled BEV neural matching-based pose solver (\texttt{DEMA}) using rasterized HD map, and then refines the estimation through a Transformer-based pose regressor (\texttt{POET}) using vectorized HD map.
The experimental results demonstrate that our method can perform robust and accurate localization under varying environmental conditions while running efficiently.
\end{abstract}

%% file: secs/1-introduction.tex
\section{Introduction}
\label{sec:intro}

\IEEEPARstart{V}{isual} localization plays a necessary role in high-level Autonomous Driving (AD) systems for its ability to provide real-time vehicle poses using an economical sensor suite.
With efforts from worldwide researchers over decades, extraordinary successes have been achieved in the visual localization area, consistently improving the accuracy, robustness, and efficiency of localization algorithms \cite{miao2024survey}.

\begin{figure}[!t]
    \centering
    \includegraphics[width=\linewidth]{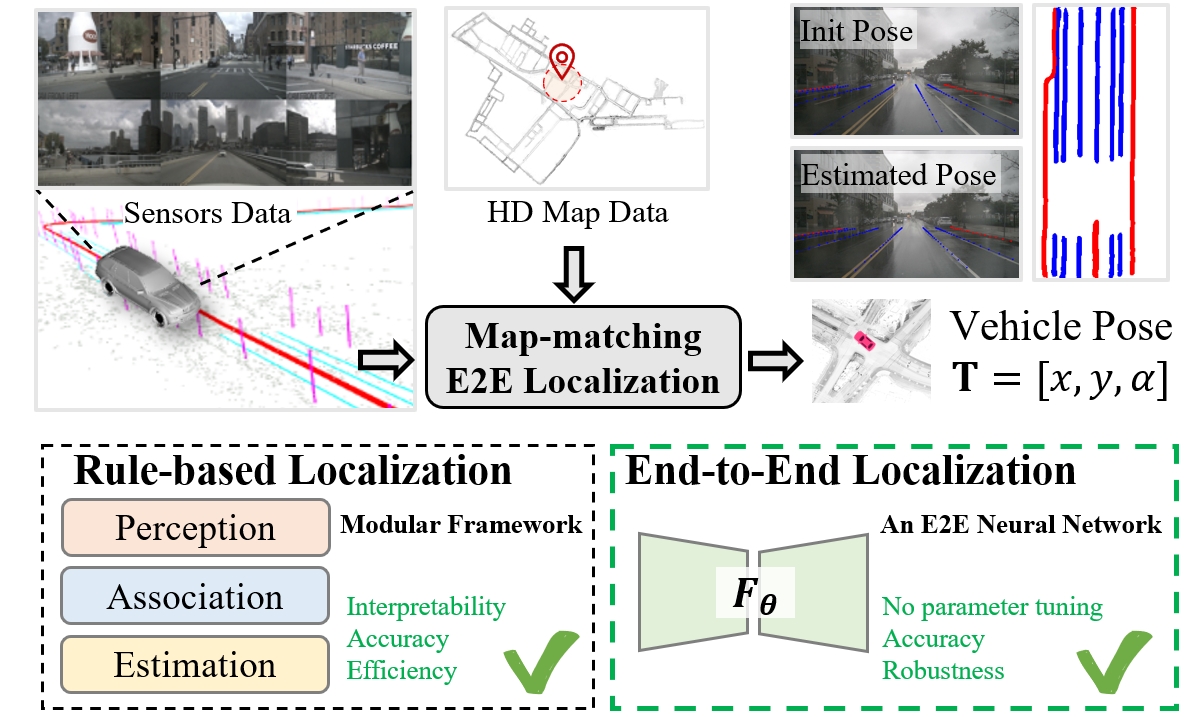}
    \caption{Traditional rule-based localization methods are built in a modular framework, involving perception, data association, and pose estimation. The end-to-end localization systems directly leverage a single neural network to estimate the vehicle pose, without the requirement of parameter tuning, while achieving better accuracy and robustness than the traditional methods.}
    \label{fig:intro}
\end{figure}

In the AD society, High-Definition (HD) maps are preferred by resource-limited Intelligent Vehicles (IVs) for it can precisely and compactly represent static driving scenarios as light-weight vectorized semantic map elements \cite{liu2020high}, greatly reducing the requirements of storage space \cite{xiao2020monocular,he2024egovm}. High-level IVs commonly perform map matching with HD maps to consistently obtain high-precision poses \cite{miao2024survey}. In general, HD map-based localization algorithms are designed as a modular pipeline \cite{xiao2020monocular,xiao2018monocular,tm3loc}, where online perception is first performed to perceive semantic elements from images, and then data association is performed between perceived map elements and HD map data, and finally pose estimation can be done on the basis of matching information, as shown in Fig. \ref{fig:intro}. Although existing solutions have already achieved promising performance in specific scenes, such a modular framework is theoretically fragile due to its inherent limitations, hindering the large-scale deployment of localization systems.

On the one hand, traditional visual localization methods rely on precise perception and data association, but cameras inherently lack depth information, which challenges the 2D-3D association between perception results and map data.
Some methods project 3D map elements into 2D image planes given pose hypotheses, and perform data association within the 2D image space \cite{xiao2018monocular,liao2020coarse}. However, the perspective effect of the camera model increases the difficulties of data association \cite{tm3loc}.
Alternatively, some other methods transform the perceived map elements into the Bird's-Eye-View (BEV) space \cite{jang2021lane,vivacqua2017self,deng2019semantic}. 
While these methods can retain the original shape of map elements, their precision and stability are affected by imprecise calibration and unavoidable vibration of cameras. 
In this work, we adopt a visual perception and data association approach in BEV space, but leverage a learning-based technique to guarantee the performance.
Moreover, to improve the robustness of visual perception against varying environmental conditions, we inject strong prior knowledge of the driving scenario from HD maps into online perception, following \cite{jiang2024pmapnet,mapex,jia2024diffmap}. Furthermore, we focus on solving the spatial misalignment problem between online perception and offline HD map. 

On the other hand, as a typical state estimation problem, accurate pose estimation remains an open topic for visual localization research. 
Traditional methods utilize a well-established non-linear optimization-based solver \cite{xiao2018monocular,xiao2020monocular,tm3loc} or particle filter-based solver \cite{liao2020coarse}. 
Zhao \textit{et al.} tried Iterative Closest Point (ICP)-based solver to estimate pose, but the performance was experimentally found to be worse than that of optimization-based solvers \cite{Zhao2024Towards}.
These traditional rule-based pose solvers generally work well, but they all require hyperparameter tuning and rule debugging based on trial-and-error, and they necessitate excellent perception and association, leading to instability when suffering from environmental interference. 
Moreover, the traditional modular framework bears the risk of information loss and error accumulation along the pipelines \cite{uniad}. The requirements of inputs by pose estimation do not guide the design of visual perception and data association yet, resulting in feature misalignment across modules.

Recent advances in end-to-end (E2E) strategies have spurred the development of E2E visual localization algorithms. These learning-based approaches utilize a unified neural networks to perceive environmental features, align them with HD maps, and estimate vehicle poses, as shown in Fig. \ref{fig:intro}, thereby eliminating the need for standalone modules and complex hand-crafted rules.
Initial E2E visual localization methods employing Standard Definition (SD) maps \cite{camiletto2024u,wu2024maplocnet} achieves only meter-level accuracy due to resolution limitations of SD maps. Subsequent approaches leveraging HD maps follow a similar framework: they extract BEV features from surrounding images and estimate poses by interacting these features with local HD map data \cite{bevlocator,he2024egovm}. Trained on extensive datasets, these methods automatically tune its network parameters, attaining decimeter-level accuracy and demonstrating strong robustness across diverse environmental conditions.
Nevertheless, existing solutions face critical limitations—either functioning as opaque black-box systems \cite{bevlocator} or demanding excessive computational resources \cite{he2024egovm}. Striking an optimal balance among efficiency, accuracy, and interpretability remains a key challenge in this emerging field.

To this end, in this paper, we propose a novel E2E vehicle localization algorithm, \textbf{\texttt{RAVE}}, using surrounding images, RAsterized and VEctorized HD map. 
In a unified network, the surrounding images are encoded into semantic BEV features, while HD map data are encoded into both rasterized features and vectorized features.
The map prior information is injected into the perceived BEV features by a proposed LOw-RAnk Flow-based prior fusion module (\textbf{\texttt{FLORA}}) for online perception enhancement.
To achieve efficient and accurate pose estimation, we propose an E2E hierarchical localization module.
In the coarse localization stage, a novel DEcoupled BEV neural MAtching-based pose solver (\textbf{\texttt{DEMA}}) is proposed in which we can solve 3 DoF vehicle poses by aligning perceived BEV features and rasterized map features in a divide-and-conquer manner.
In the fine localization stage, we design a Transformer-based POse rEgressor (\textbf{\texttt{POET}}) to further refine the estimated poses with vectorized map features.
The overall network can be E2E optimized, eliminating hand-crafted rules design and hyperparameter tuning.
The primary contributions are summarized as follows:

\begin{itemize}
    \item The E2E visual localization network \textbf{\texttt{RAVE}} is elaborated that performs efficient, accurate, and interpretable pose estimation with lightweight HD maps.

    \item The \textbf{\texttt{FLORA}} module is proposed to spatially align the perceived BEV feature and HD map, and enhance the online perception by fusing the prior from offline map.
    
    \item A hierarchical localization module is proposed that utilizes rasterized and vectorized map features. First, the \textbf{\texttt{DEMA}} module is proposed to solve 3 DoF poses individually, reducing sampling complexity in neural matching from cubic level to linear level. Second, the \textbf{\texttt{POET}} module is developed to refine estimated poses.

\end{itemize}

A preliminary version of this work is presented in our conference paper \cite{miao2025efficient}. 
In \cite{miao2025efficient}, the \textbf{\texttt{DEMA}} solver is proposed to efficiently solve pose in an E2E localization network.
In the current paper, we first design the \textbf{\texttt{FLORA}} module to incorporate the misaligned map prior to online BEV perception. Then, we substantially extend the E2E localization network by proposing the \textbf{\texttt{POET}} module, further improving the localization accuracy in the hierarchical localization scheme. Finally, we extend the experimental evaluation of our approach and obtain more comprehensive and reliable conclusions.

%% file: secs/2-relatedwork.tex
\section{Related Works}
\label{review}

\subsection{Learning-based Online HD Mapping}

Recently, constructing HD map elements online through a BEV perception framework has been increasingly adopted by the community as it reduces the substantial human effort required by conventional rule-based mapping pipelines \cite{onlinemappingsurvey}.
HDMapNet \cite{hdmapnet} builds the first online mapping framework that transforms surrounding images into BEV features by a learnable neural network, fuses cross-modal features and directly learns map elements in BEV space.
SuperFusion \cite{dong2024superfusion} improves the long-range mapping performance by fusing visual and LiDAR data at multiple levels. 
In contrast to these two methods perceiving rasterized map, VectorMapNet \cite{liu2023vectormapnet} directly estimates vectorized HD map elements through a Transformer decoder architecture \cite{detr}. 
This pipeline facilitates the perception of intricate and detailed shapes of map elements.
MapTR series \cite{liao2022maptr,liao2024maptrv2} propose a permutation-equivalent shape modeling method for unified representations of map elements.
BeMapNet \cite{bemapnet} models semantic map elements as piecewise Bezier curves and achieves excellent perception performance.
These works all provide a promising solution for converting raw sensor inputs into BEV features that share the same representation space with HD map, paving the way for data association in subsequent E2E vehicle localization.

\subsection{Map Prior-enhanced BEV Perception}

Due to occlusion, limited perception field, and varying environmental conditions, the online perception is theoretically challenging to achieve stable performance even with sophisticated algorithms. 
Therefore, many studies have incorporated the prior knowledge from offline map data into online BEV perception \cite{onlinemappingsurvey}.
Some methods learn structural prior of HD map in its offline training stage. P-MapNet \cite{jiang2024pmapnet} utilizes masked autoencoder to learn the distribution of HD map elements so that the imprecise perception can be corrected. DiffMap \cite{jia2024diffmap}, instead, introduces a diffusion module for sampling purposes and it formulates the perception enhancement as a denoising process. Some recent methods enhance the mapping performance through incorporation with prior maps in the online perception stage, such as SD maps \cite{jiang2024pmapnet,zeng2024driving}, outdated HD maps \cite{mapex,zeng2024driving}, and historically perceived maps \cite{nmp,zeng2024driving}. 
P-MapNet \cite{jiang2024pmapnet} integrates SD map features using cross attention mechanism. 
NMP \cite{nmp} builds a global neural map prior during historical inference stage and injects the historical prior by a Gated Recurrent Unit (GRU)-based module in the online perception stage. PriorDrive \cite{zeng2024driving} flexibly integrates different prior map in both BEV features and learnable queries of the map decoder with a unified framework.
In localization tasks where HD maps exist, we can enhance the online mapping performance with offline HD map. However, existing studies ideally assume precise vehicle poses are given. Our work addresses this limitation by considering the spatial alignment between online perception and HD map.

\subsection{Visual Localization based on HD Map}

Traditional HD map-based localization systems are generally built in a modular framework \cite{miao2024survey}. They first extract semantic map elements (\textit{e.g.}, lanes \cite{pink2008visual}, poles \cite{spangenberg2016pole}, traffic signs \cite{qu2015vehicle}) from images, then perform map element matching with the HD map, and finally solve poses based on the matching information.
However, indistinguishable elements, missed and false detections \cite{Jiang2023Data}, and inaccurate view transformations \cite{Wan2024Monocular} make the rule-based map matching process highly ambiguous. 
Moreover, traditional pose solvers including both non-linear optimizers \cite{xiao2018monocular,Wan2024Monocular,tm3loc,xiao2020monocular} and particle filters \cite{liao2020coarse} heavily rely on precise matching information. 
Low-quality data association results will result in wrong convergence in pose estimation \cite{Jiang2023Data}. Such a modular framework is fragile and the rule-based solution requires costly hyperparameter tuning, blocking the wide usage in varying driving scenarios.

Recently, researchers have begun to develop learning-based localization methods using HD map data. 
Zhang \textit{et al.} employed DNN models to directly estimate relative longitudinal and latitudinal positions between local semantic map and HD map \cite{zhang2022learning}, but the solution requires multiple sensor fusion to build the local semantic map. 
Zhao \textit{et al.} utilized BEV perception network to obtain BEV map elements and estimated pose by traditional optimization-based solver \cite{Zhao2024Towards}.
In pursuit of a fully E2E system that directly estimates vehicle poses from raw sensory data, BEVLocator \cite{bevlocator} employs a Transformer-based network to jointly perform feature extraction, data association, and pose estimation. It achieves $<$20cm localization accuracy in longitudinal and lateral position, and it requires only single-frame surrounding images, as opposed to sequential images \cite{tm3loc,xiao2020monocular}. 
Subsequently, EgoVM \cite{he2024egovm} enhances localization accuracy and interpretability by using more informative HD maps and a neural matching-based pose solver. 
OrienterNet \cite{sarlin2023orienternet} also utilizes neural matching to estimate pose by matching SD map with monocular image. 
SegLocNet \cite{zhou2025seglocnet} achieves further improvements by fusing surrounding images and LiDAR. 
These works first transform image features into BEV space, then estimate poses based on BEV features and map features by learnable network, reducing the requirements of hand-crafted rules and hyperparameter tuning, which motivates our work. These solutions achieve great robustness and localization accuracy,  but they are difficult to achieve a balance among accuracy, efficiency, and interpretability, limiting the application to high-level AD tasks requiring high safety.

%% file: secs/3-method.tex
\section{Methodology}
\label{sec:method}

In this section, we first define the visual-to-HD map localization problem using surrounding cameras. Next, we present the overall framework of the proposed E2E localization network. Then, we describe the four core modules in the network, \textit{i.e.}, BEV perception backbone, HD map encoder, prior fusion module, and localization module in detail. Finally, the training process of the proposed network and an adaptive localization scheme are given.

\subsection{Problem Formulation}
\label{sec:definitation}

Following a common assumption in the AD community that the ground vehicle is attached to the flatten ground, the considered vehicle pose in the global frame can be degraded to 3 DoF in BEV space, \textit{i.e.}, $\textbf{T}=\left[x, y, \alpha \right] \in \mathbb{SE}(2)$, where $x$ and $y$ denote the longitudinal and lateral distance between the origins of the ego frame and the global frame, and $\alpha$ is the heading/yaw angle. 
Generally, a HD map is composed of vectorized semantic map elements, $\mathcal{M}=\{\textbf{m}_i\}$, where a map element (such as lane divider and pedestrian crossing) can be represented by a set of 2D points in the global frame with semantic category.

Given an initial vehicle pose $\textbf{T}_0$ that is commonly obtained by GNSS signals, the HD map utilized in visual-to-HD map localization tasks is transformed to a virtual viewpoint on the initial vehicle pose and then cropped to a local map $\mathcal{M}_0$: 
\begin{equation}
    \mathcal{M}_0=\texttt{crop}(\texttt{transform}(\mathcal{M}, \textbf{T}_0))
\end{equation}
Therefore, the visual-to-HD map localization aims to estimate an optimal vehicle pose $\triangle\hat{\textbf{T}} =\left[\triangle \hat{x}, \triangle \hat{y}, \triangle \hat{\alpha} \right] \in \mathbb{SE}(2)$, which aligns the surrounding images $\mathcal{I}$ to the local HD map $\mathcal{M}_0$:
\begin{equation}
    \triangle\hat{\textbf{T}}=\mathop{\arg\min}\limits_{\triangle\textbf{T}} d(\mathcal{I}, \texttt{transform}(\mathcal{M}_0, \triangle\textbf{T}))
\end{equation}
where $d(\cdot,\cdot)$ measures the differences between cross-modal data. After that, the vehicle pose can obtained by a pose transformation process, $\hat{\textbf{T}}=\textbf{T}_0 \otimes \triangle\hat{\textbf{T}}$:
\begin{equation}
    \begin{aligned}
        \hat{x} &= x_0 + \triangle \hat{x}\cdot\texttt{cos}\alpha_0-\triangle \hat{y}\cdot\texttt{sin}\alpha_0 \\
        \hat{y} &= y_0 + \triangle \hat{x}\cdot\texttt{sin}\alpha_0+\triangle \hat{y}\cdot\texttt{cos}\alpha_0 \\
        \hat{\alpha} &= \alpha_0 + \triangle \hat{\alpha}
    \end{aligned}
\end{equation}

\subsection{Overview of the E2E Localization Network}
\label{sec:overview}

\begin{figure*}[!t]
    \centering
    \includegraphics[width=0.97\linewidth]{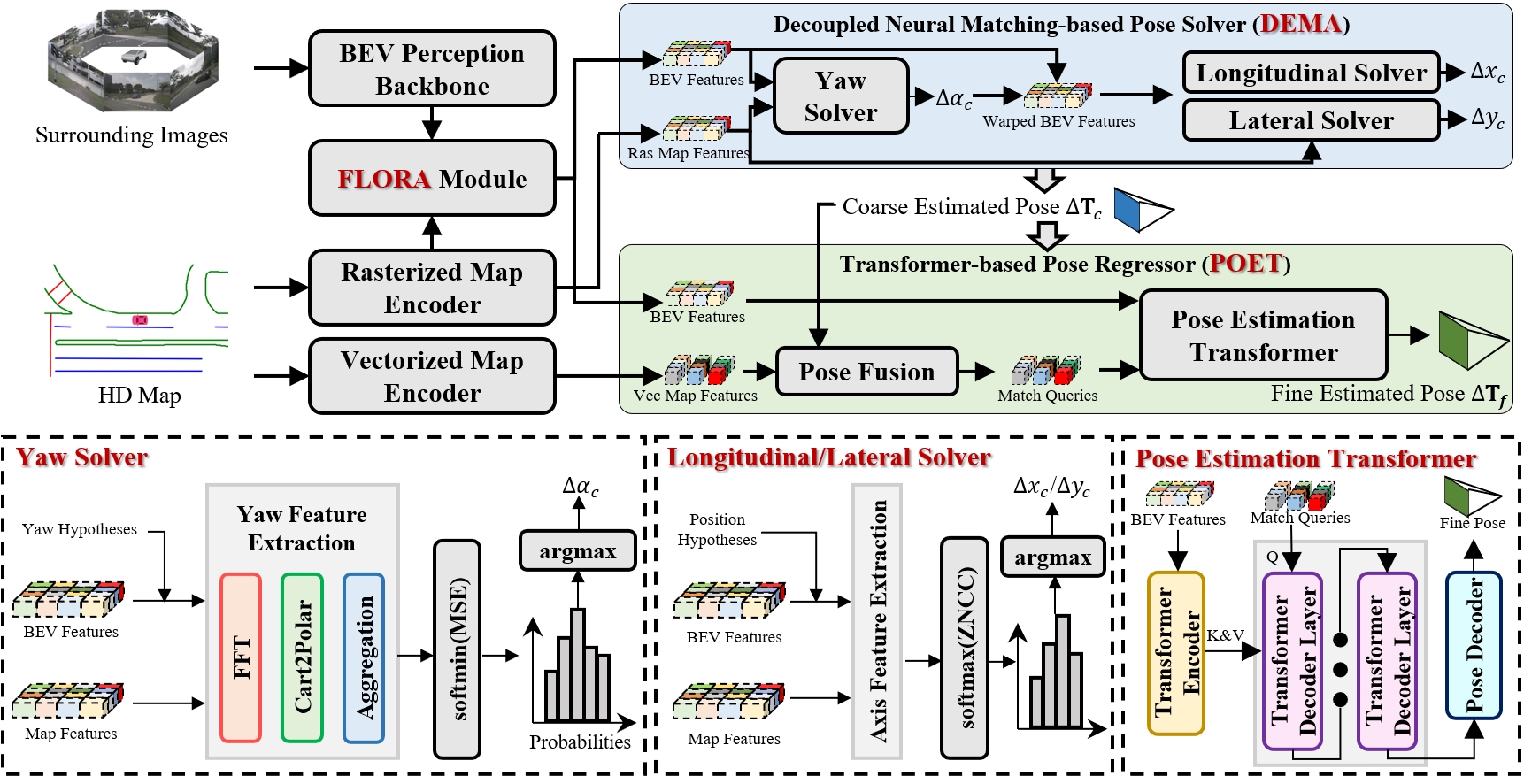}
    \caption{The overall framework of the proposed end-to-end localization network \textbf{\texttt{RAVE}}.}
    \label{fig:framework}
\end{figure*}

To perform HD map-based localization in an E2E manner, the proposed method incorporates both BEV perception and localization in a unified network. As shown in Fig. \ref{fig:framework}, the network is fed by the surrounding images $\mathcal{I}$ and the local HD map $\mathcal{M}_0$. 
We adopt an online mapping network as the visual backbone to obtain BEV features $\textbf{F}_\mathcal{I}$ from surrounding images. 
Meanwhile, the local HD map is represented both by a rasterized semantic mask $\textbf{M}$ in the BEV space and by a set of vectorized map embeddings $\textbf{E}$.
We extract rasterized map features $\textbf{F}^R_\mathcal{M}$ from $\textbf{M}$ by a rasterized map encoder network and vectorized map features $\textbf{F}^V_\mathcal{M}$ from $\textbf{E}$ through a vectorized map encoder, respectively. 
Then, $\textbf{F}^R_\mathcal{M}$ are utilized to enhance the perception quality of $\textbf{F}_\mathcal{I}$ as a strong prior via the proposed \textbf{\texttt{FLORA}} module.
Finally, a hierarchical end-to-end localization module is performed to accurately and efficiently estimate vehicle pose $\triangle\hat{\textbf{T}}_f$, which first estimates coarse pose $\triangle\hat{\textbf{T}}_c$ by utilizing $\textbf{F}^R_\mathcal{M}$ in the proposed \textbf{\texttt{DEMA}} solver, and then refines the estimation with $\textbf{F}^V_\mathcal{M}$ in the proposed \textbf{\texttt{POET}} module.

\subsection{BEV Perception Backbone and HD Map Encoder}
\label{sec:backbone}

The surrounding images and HD map data are in different modalities, precise cross-modal matching and pose estimation between them is technically challenging for traditional localization systems. 
To enable cross-modal matching, we extract high-dimensional learnable features from both surrounding images and HD map data.

First, we employ a well-established online mapping network, BeMapNet \cite{bemapnet}, as the BEV perception backbone. The backbone takes a set of surrounding images $\mathcal{I}$, intrinsics $\mathcal{K}$ and extrinsics $\mathcal{U}$ of cameras as input, and extracts Perspective-View (PV) features from each image. Then, the backbone transforms and aggregates PV features into BEV space by a Transformer-based view transformation module, and finally obtains BEV semantic features $\textbf{F}_\mathcal{I}$. We use a simple convolution segmentation head $\mathcal{H}$ to decode $\textbf{F}_\mathcal{I}$ to a semantic mask $\hat{\textbf{M}}_\mathcal{I} \in {(0,1)}^{H \times W \times (C+1)}$, where $H,W$ are the sizes of the map mask, and $C$ is the number of semantic categories:
\begin{align}
    \textbf{F}_\mathcal{I} &= \texttt{imgbackbone}(\mathcal{I,K,U})\\
    \hat{\textbf{M}}_\mathcal{I} &=\mathcal{H}(\textbf{F}_\mathcal{I})
\end{align}
Since the map elements are sparse in the HD map, we add a background class in the semantic mask to make the BEV features informative for subsequent E2E localization.

Meanwhile, we build a VGG-like network \cite{vgg} as the rasterized map encoder to extract rasterized map features $\textbf{F}^R_\mathcal{M}$ from rasterized HD map mask $\textbf{M} \in {\{0,1\}}^{H \times W \times (C+1)}$.
We also reconstruct the map mask by $\textbf{F}^R_\mathcal{M}$ so that the map features can maintain semantic and spatial information of the HD map.
\begin{align}
    \textbf{F}^R_\mathcal{M} &= \texttt{encoder}_\texttt{ras}(\textbf{M})\\
    \hat{\textbf{M}}_\mathcal{M} &=\mathcal{H}(\textbf{F}^R_\mathcal{M})
\end{align}
where $\hat{\textbf{M}}_\mathcal{M}$ is the reconstructed HD map mask.

Additionally, we extract fine-grained map features from vectorized HD map data.
Concretely, the HD map data is represented by element-wise hybrid embeddings. For each map element $\textbf{m}_i$, the 2D positions and direction of its map points are embedded by Fourier position encoding \cite{tancik2020fourier}:
\begin{align}
    \textbf{E}^{pos}_i &= \texttt{Fourier}([u^0_i,u^1_i,...])\\
    \textbf{E}^{dir}_i &= \texttt{Fourier}([v^0_i,v^1_i,...])
\end{align}
where $u^k_i$ is the 2D position of the $k$-th point in the $\textbf{m}_i$ and $v^k_i=u^{k+1}_i-u^{k}_i$ is the direction between adjacent points in the $\textbf{m}_i$. The semantic category $c_i$ and the index of the element $i$ are embedded by learnable embeddings, $\textbf{E}^{sem}_i, \textbf{E}^{ins}_i$, respectively. Thus, a map element is embedded as a fixed-size structured embedding by concatenating all the map information:
\begin{equation}
    \textbf{E}_i=[\textbf{E}^{pos}_i,\textbf{E}^{dir}_i,\textbf{E}^{sem}_i,\textbf{E}^{ins}_i] \in \mathbb{R}^{N_p \times D}
\end{equation}
where $N_P$ is the desirable amount of map points in each map element and $D$ is the feature dimension.
For those elements without sufficient map points, we pad the elements with 0. Finally, the vectorized map features $\textbf{F}^V_\mathcal{M}$ are obtained by aggregating map embeddings within each map elements through a vectorized encoder \cite{gao2020vectornet}:
\begin{equation}
    \textbf{F}^V_\mathcal{M} = \texttt{encoder}_\texttt{vec}({\textbf{E}})
\end{equation}

By converting raw surrounding images and HD map data into learnable features, the difficulties of cross-modal data association, fusion, and pose estimation can be effectively addressed by E2E learning.

\subsection{FLORA: Low-rank Flow-based Prior Fusion Module}
\label{sec:flora}

\begin{figure}[!t]
    \centering
    \includegraphics[width=0.97\linewidth]{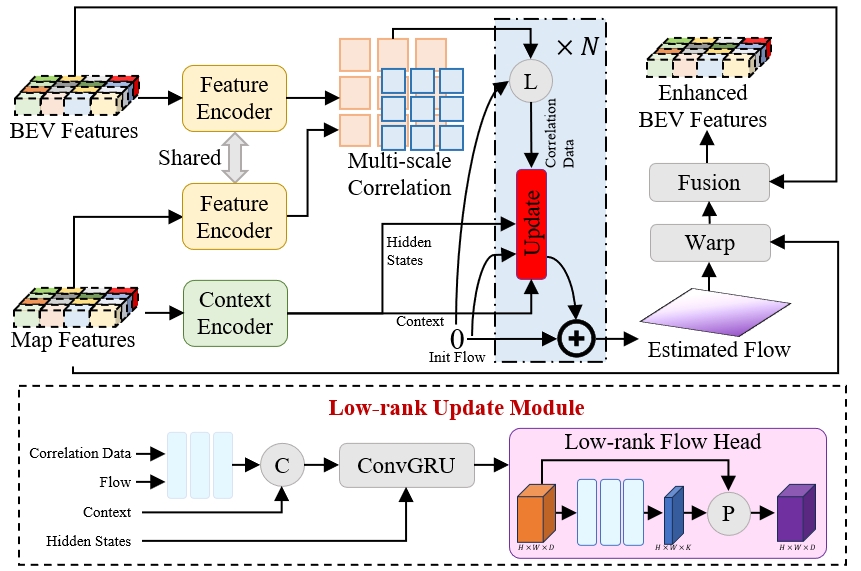}
    \caption{The detailed structure of the proposed prior fusion network, \textbf{\texttt{FLORA}}.}
    \label{fig:flow}
\end{figure}

As the visual measurements fed into the E2E localization framework, the BEV features from surrounding cameras play an important role in localization stability and accuracy. 
In this work, the BEV features are trained to encode static map element information, which shares the same feature space with HD map data. 
Thus, the offline HD map will be a strong prior for online BEV perception, which motivates several map-enhanced perception works \cite{jia2024diffmap,jiang2024pmapnet}. 
However, without high-precision vehicle pose, there is a spatial displacement between BEV perception results and the retrieved local HD map $\mathcal{M}_0$, which affects the perception enhancement process. 
Therefore, we propose a flow-based prior fusion module, \textbf{\texttt{FLORA}}, which first corrects the spatial displacements between BEV features and rasterized map features based on estimated flows, and then fuses them to achieve enhanced BEV perception.

In order to estimate the pixel-level displacement between $\textbf{F}_\mathcal{I}$ and $\textbf{F}^R_\mathcal{M}$, we build a light-weight flow estimation network, as shown in Fig. \ref{fig:flow}. Similar to \cite{raft}, a shared feature encoder is utilized to extract features from both $\textbf{F}_\mathcal{I}$ and $\textbf{F}^R_\mathcal{M}$, and then a multi-scale local correlation volume is obtained by calculating the pixel-wise similarities between neighboring pixels across two features. Since $\textbf{F}^R_\mathcal{M}$ is more stable and accurate, it is selected to generate the initial hidden states and the context feature through a context encoder. Under an iterative refinement framework, the network retrieves corresponding correlation data from correlation volume based on estimated flow from the previous iteration, and updates the estimated flow through an update module. In the update module, the correlation data, previously estimated flow, context feature, and previous hidden states are aggregated and updated, resulting in a flow feature, which is finally decoded to BEV flow.

{
\setlength{\parindent}{0cm}
\textbf{Low-rank BEV Flow:}
In the proposed \textbf{\texttt{FLORA}} module, the update module involves a low-rank feature projection process. Given that the spatial displacement between BEV features and map data is caused by 3 DoF pose errors, the displacement in the BEV space should be low-rank and regular. Thus, inspired by \cite{basehomo}, we reduce the rank of the flow feature so that the dominant flow feature can be retained while noise will be rejected. Formally, the flow feature $\textbf{F}_f\in \mathbb{R}^{H_f\times W_f \times D}$ is compressed in the channel dimension to reduce its rank, resulting in a feature $\textbf{F}_v\in \mathbb{R}^{H_f\times W_f \times K}$ with $K$ channels. Each channel of $\textbf{F}_v$ can serve as a flow feature basis $v_k \in \mathbb{R}^{H_fW_f}, k=1,2,...,K$ after flattening. Finally, the original flow feature $\textbf{F}_f$ is projected into the subspace of the flow feature bases so that we obtain a low-rank flow feature $\textbf{F}^{l}_f \in \mathbb{R}^{H_f\times W_f \times D}$:
}
\begin{equation}
    \textbf{F}^l_f=V{(V^TV)}^{-1}V^T\cdot \textbf{F}_f
\end{equation}
where $V=[v_1,...,v_{K}]$. The normalization term ${(V^TV)}^{-1}$ is required since ${v_k}$ is not guaranteed to be orthogonal. 

With the estimated flow from the \textbf{\texttt{FLORA}} module, the rasterized map feature $\textbf{F}^R_\mathcal{M}$ is warped so that the spatial displacement is effectively reduced. The warped map feature and the BEV feature are spatially aligned. Therefore, we simply concatenate these two features in channel dimension and fuse them by a convolution layer. The overall procedure in the proposed \textbf{\texttt{FLORA}} module can be formatted as:
\begin{equation}
    \textbf{F}^e_\mathcal{I}=\textbf{\texttt{FLORA}}(\textbf{F}_\mathcal{I},\textbf{F}^R_\mathcal{M})=\texttt{conv}([\textbf{F}_\mathcal{I}, \texttt{flow}(\textbf{F}_\mathcal{I},\textbf{F}^R_\mathcal{M})\circledast \textbf{F}^R_\mathcal{M}])
\end{equation}
where $\circledast$ denotes the warp operator based on estimated flow.

By fusing strong prior information from the HD map feature, the quality of BEV perception feature is significantly improved. We also decode the enhanced BEV features $\textbf{F}^e_\mathcal{I}$ to map mask for training:
\begin{equation}
    \hat{\textbf{M}}^e_\mathcal{I} =\mathcal{H}(\textbf{F}^e_\mathcal{I})
\end{equation}

\subsection{Hierarchical End-to-end Localization Module}
\label{sec:localization}

After obtaining enhanced high-quality BEV features $\textbf{F}^e_\mathcal{I}$, rasterized map features $\textbf{F}^R_\mathcal{M}$, and vectorized map features $\textbf{F}^V_\mathcal{M}$, we propose to estimate vehicle pose relative to the local HD map $\mathcal{M}_0$ in the coarse-to-fine localization framework, where we utilize rasterized HD map features in the coarse stage and vectorized HD map features in the fine stage.

\subsubsection{Decoupled BEV Neural Matching-based Coarse Localization using Rasterized Map}
Considering that the spatial displacement between BEV perception and HD map is caused by inaccurate 3 DoF vehicle pose, we first utilize a decoupled neural matching-based solver \textbf{\texttt{DEMA}} for interpretable and efficient coarse pose estimation using $\textbf{F}^e_\mathcal{I}$ and $\textbf{F}^R_\mathcal{M}$:
\begin{equation}
    \triangle \hat{\textbf{T}}_c=\textbf{\texttt{DEMA}}(\textbf{F}^e_\mathcal{I},\textbf{F}^R_\mathcal{M})
\end{equation}

Note that we compress the channel dimension and size of both $\textbf{F}^e_\mathcal{I}$ and $\textbf{F}^R_\mathcal{M}$ before feeding them into the \textbf{\texttt{DEMA}} module for computational efficiency, \textit{i.e.}, $\textbf{F}^e_\mathcal{I},\textbf{F}^R_\mathcal{M} \in \mathbb{R}^{H_c\times W_c\times D_c}$. 

{
\setlength{\parindent}{0cm}
\textbf{Preliminary: Full Neural Matching (\textbf{\texttt{FUMA}})}
Traditional \textbf{\texttt{FUMA}} solver in BEV space exhaustively samples all possible 3 DoF pose hypotheses by grid sampling: $\{\triangle\textbf{T}_{ijk}=[\triangle x_i,\triangle y_j,\triangle \alpha_k]\}$, where $i \in [1, N_x], j \in [1, N_y], k \in [1, N_\alpha]$ and $N_x,N_y,N_\alpha$ are the numbers of hypotheses sampled in each dimension, and then estimates their probabilities based on the feature similarities \cite{weston2022fast,he2024egovm,bevpose}. 
Theoretically, the amount of pose hypotheses in the \textbf{\texttt{FUMA}} solver is as high as $N_x \times N_y \times N_\alpha$, leading to unacceptable computational burden when the hypothesis space is extremely large. 
}

{
\setlength{\parindent}{0cm}
\textbf{Decoupled Neural Matching (\textbf{\texttt{DEMA}})}
In this paper, we propose to decouple the feature representation affected by each DoF in pose estimation and estimate each DoF through the \textbf{\texttt{DEMA}} solver in a divide-and-conquer framework. 
As shown in Fig. \ref{fig:framework}, the proposed pose solver first extracts \textit{yaw feature} from $\textbf{F}^e_\mathcal{I}$ and $\textbf{F}^R_\mathcal{M}$ and estimates the yaw angle by measuring differences between \textit{yaw features}. Second, $\textbf{F}^e_\mathcal{I}$ is transformed to $\textbf{F}^\alpha_\mathcal{I}$ by estimated yaw angle $\triangle \hat{\alpha}$ so that the $\textbf{F}^\alpha_\mathcal{I}$ and $\textbf{F}^R_\mathcal{M}$ differ only in longitudinal and lateral positions. Finally, we extract \textit{axis feature} from $\textbf{F}^\alpha_\mathcal{I}$ and $\textbf{F}^R_\mathcal{M}$ and solve longitudinal and lateral positions. Detailed descriptions are given below.
}

\begin{figure}[!t]
    \centering
    \includegraphics[width=0.97\linewidth]{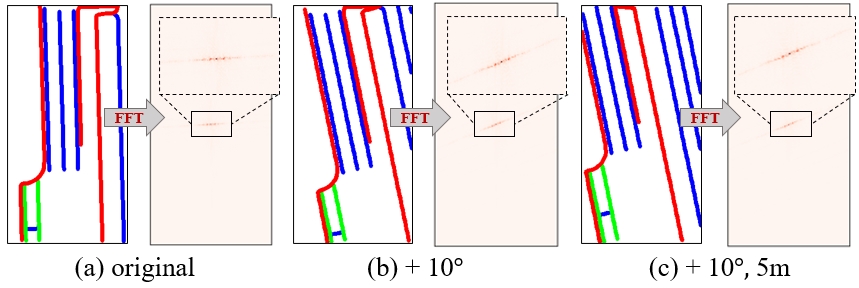}
    \caption{Visualization cases of the amplitude-frequency features for map mask. The left image in each group is HD map mask while the right one is its amplitude-frequency graph. (a) shows an original HD map, (b) shows a map after adding 10$\degree$  yaw deviation, and (c) shows a map after adding 10$\degree$ yaw deviation and 5m deviation in both longitudinal and lateral directions.}
    \label{fig:freq}
\end{figure}

{
\setlength{\parindent}{0cm}
\textbf{Yaw Solver by Amplitude-Frequency Features}
As shown in Fig. \ref{fig:freq}, the amplitude-frequency graph of an image is only affected by its orientation but not its position \cite{weston2022fast,FreSCo}. 
Therefore, the yaw solver of the \textbf{\texttt{DEMA}} module individually solves yaw angle by decoupling the BEV feature representation affected by yaw angle. 
Formally, we generate yaw hypotheses by grid searching: $\{\triangle \alpha_k | 1\leq k \leq N_\alpha\}$ and transform $\textbf{F}^e_\mathcal{I}$ accordingly, resulting in transformed BEV features $\{{\textbf{F}}^{\alpha(k)}_\mathcal{I} | 1\leq k \leq N_\alpha\}$. 
Then, $\textbf{F}^{\alpha(k)}_\mathcal{I}$ and $\textbf{F}^R_\mathcal{M}$ are converted into frequency domain by Fast Fourier Transform (\texttt{FFT}) process to obtain Frequency-BEV (F-BEV) features. 
To quantitatively represent the amplitude-frequency features affected by yaw angle along the yaw axis in polar coordinates, we transform the F-BEV features in Cartesian coordinates into Polar F-BEV (PF-BEV) features in polar coordinates. Finally, we aggregate the PF-BEV features perpendicular to yaw axis and obtain corresponding \textit{yaw features}:
}
\begin{equation}
    \textbf{Y}_*=\texttt{sum}(\texttt{Cart2Polar}(\texttt{FFT}(\textbf{F}_*)))
\end{equation}
The probability of each yaw hypothesis is calculated by normalized Mean Square Error (\texttt{MSE}) between \textit{yaw features}:
\begin{equation}
    p(\triangle \alpha_k)=\texttt{softmin}(\texttt{MSE}(\textbf{Y}_\mathcal{M},\textbf{Y}^{\alpha(k)}_\mathcal{I}))
\end{equation}
and an optimal yaw angle $\triangle \hat{\alpha}$ can be estimated:
\begin{equation}
    \label{eqn:yaw}
    \triangle \hat{\alpha}_c=\triangle\alpha_{k},~k=\mathop{\arg\max}\limits_{k} p(\triangle \alpha_k)
\end{equation}

The estimated yaw angle is utilized to transform $\textbf{F}^e_\mathcal{I}$, correcting the orientation differences between features.

\begin{figure}[!t]
    \centering
    \includegraphics[width=0.97\linewidth]{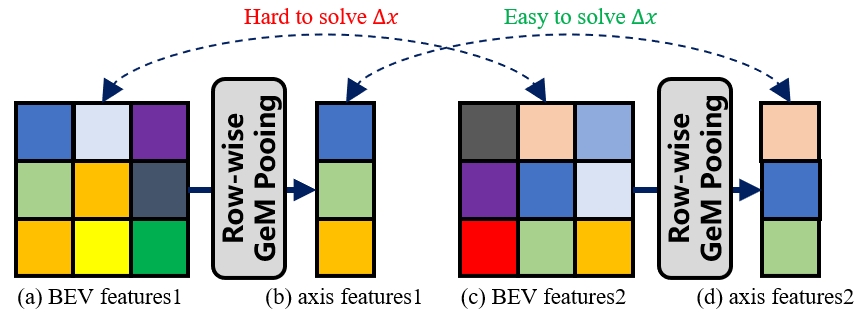}
    \caption{Diagram of \textit{axis feature} extraction by GeM pooling \cite{gempooling}. (a) and (c) show two original BEV features whereas (b) and (d) are their corresponding \textit{axis features}.}
    \label{fig:axis}
\end{figure}

{
\setlength{\parindent}{0cm}
\textbf{Longitudinal/Lateral Solver by Axis Features}
The position differences of features in longitudinal dimension can be quantitatively reflected in the row order of features, as shown in Fig. \ref{fig:axis} (a) and (c), but the position difference in the other dimension (lateral) will make this phenomenon indistinct. Therefore, to individually solve longitudinal and lateral positions, we propose to aggregate features along the single dimension in the longitudinal and lateral solvers so that the aggregated \textit{axis features} differ only in the position of one dimension, as shown in Fig. \ref{fig:axis} (b) and (d). For instance, to solve longitudinal position $\triangle x$, the longitudinal solver samples longitudinal hypotheses by grid sampling: $\{\triangle x_i | 1\leq i \leq N_x\}$ and transform $\textbf{F}^{e}_\mathcal{I}$ into features $\{{\textbf{F}}^{x(i)}_\mathcal{I} | 1\leq i \leq N_x\}$.
}
The transformed BEV features and map feature are aggregated by GeM pooling \cite{gempooling} along a single row dimension and generate corresponding \textit{axis feature} $\textbf{A}_* \in\mathbb{R}^{H_c \times D_c}$:
\begin{equation}
    \textbf{A}_*[u] = (\frac{1}{W_c}\sum^{W_c}_{v=1}(\textbf{F}[u,v])^{\omega})^{1/\omega}
\end{equation}
where $u,v$ are the row and column index, $\omega$ is the learnable weighting factor. Then, the probability of each longitudinal hypothesis is the normalized Zero-mean Normalized Cross-Correlation (\texttt{ZNCC}) between \textit{axis features}:
\begin{equation}
    p(\triangle x_i)=\texttt{softmax}(\texttt{ZNCC}(\textbf{A}_\mathcal{M},\textbf{A}^{x(i)}_\mathcal{I}))
\end{equation}
and an optimal longitudinal position $\triangle \hat{x}_c$ is estimated. The lateral solver performs in a technically similar way but it aggregates features along a single column dimension. Finally, we obtain a coarse vehicle estimation $\triangle \hat{\textbf{T}}_c =[\triangle \hat{\alpha}_c, \triangle \hat{x}_c, \triangle \hat{y}_c]$.

\subsubsection{Transformer-based Fine Localization using Vectorized Map}
Although the \textbf{DEMA} solver provides efficient pose estimation in coarse localization, its localization accuracy is still limited, especially in the longitudinal pose estimation.
We attribute this limitation to the resolution constraints of BEV perception and map encoding, which restrict the granularity of input features fed into the pose estimation module.
In order to achieve localization with higher accuracy and correct unreliable pose estimation from the coarse localization module, we propose a Transformer-based pose regressor \texttt{\textbf{POET}} to refine pose estimation using enhanced BEV features $\textbf{F}^e_\mathcal{I}$, vectorized map features $\textbf{F}^V_\mathcal{M}$, and coarse pose estimation $\triangle \hat{\textbf{T}}_c$ with corresponding probability $\hat{\textbf{P}}_c=\{p(\triangle \hat{x}_c), p(\triangle \hat{y}_c), p(\triangle \hat{\alpha}_c)\}$:
\begin{equation}
\label{eqn:fine}
    \triangle \hat{\textbf{T}}_f=\texttt{\textbf{POET}}(\textbf{F}^e_\mathcal{I},\textbf{F}^V_\mathcal{M}|\triangle \hat{\textbf{T}}_c,\hat{\textbf{P}}_c)
\end{equation}

{
\setlength{\parindent}{0cm}
\textbf{Preliminaries: Transformer attention \cite{vaswani2017attention}}
We apply a Transformer in this work for pose estimation. For better readability, we briefly review its core component, attention layer, as background. Attention layers take $d$-dimensional query vector $q$, key vector $k$, and value vector $v$ and conduct the following calculation:
}
\begin{equation}
\label{equ:attn}
    \texttt{attn}(q,k,v)=\texttt{softmax}(\frac{qk^T}{\sqrt{d}})v
\end{equation}
Intuitively, $q$ retrieves related information from $v$ based on the relevance between $q$ and $k$. Then, the retrieved values are used to update $q$ in the Transformer layer. 

\begin{figure}[!t]
    \centering
    \includegraphics[width=0.97\linewidth]{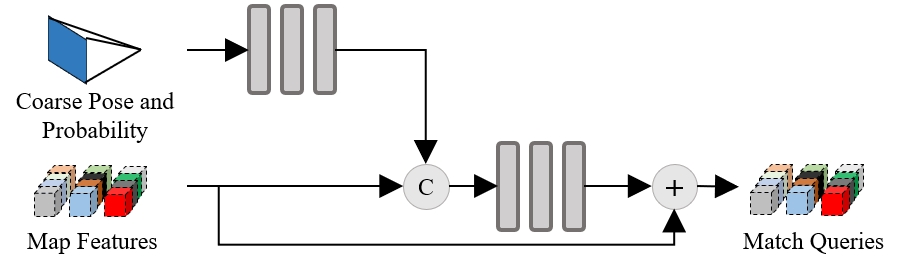}
    \caption{The network structure of the pose fusion module in the proposed \textbf{\texttt{POET}} module.}
    \label{fig:pose_fusion}
\end{figure}

In order to make Transformer-based neural network applicable in map matching-based localization tasks, we first initialize the queries for Transformer based on the vectorized map features $\textbf{F}^V_\mathcal{M}$ following \cite{bevlocator}, but we improve this by incorporating the coarse pose estimation as well.
As shown in Fig. \ref{fig:pose_fusion}, the pose estimation from the coarse localization module is utilized to fuse with $\textbf{F}^V_\mathcal{M}$ in a pose fusion module so that the built queries initially contain coarse pose information if the coarse estimation can be referenced. 
Concretely, the estimated coarse pose $\triangle \hat{\textbf{T}}_c$ and probability $\hat{\textbf{P}}_c$ are both converted to high-dimensional features via a simple Multiple Layer Perceptron (\texttt{MLP}) module. then concatenated with $\textbf{F}^V_\mathcal{M}$ in the channel dimension and fused through another MLP with residual connection, generating queries $Q_T$ containing referenced information from the coarse pose estimation, dubbed as \textit{match queries}:
\begin{equation}
    Q_T = \texttt{MLP}([\textbf{F}^V_\mathcal{M},\texttt{MLP}([\triangle \hat{\textbf{T}}_c,\hat{\textbf{P}}_c])])+\textbf{F}^V_\mathcal{M}
\end{equation}

Then, as shown in Fig. \ref{fig:framework}, the enhanced BEV features $\textbf{F}^e_\mathcal{I}$ are first processed by a Transformer encoder to further enhance their representative quality:
\begin{equation}
    \textbf{F}^T_\mathcal{I} = \texttt{selfattn}(q=k=v=\textbf{F}^e_\mathcal{I})
\end{equation}
The \textit{match queries} $Q_T$ are utilized as the queries in multiple Transformer decoder layers to iteratively interact with BEV features (keys and values):
\begin{equation}
    Q_T = \texttt{crossattn}(q=Q_T,k=v=\textbf{F}^T_\mathcal{I})
\end{equation}
In each iteration, the \textit{match queries} are refined by the relevant information from BEV features, which mimic the data association process in the traditional map matching-based localization but act in an implicit way. 
After several iterations, the updated \textit{match queries} contain the implicit matching information between BEV features and map data, which can be decoded to the fine vehicle pose $\triangle \textbf{T}_f$ with high-precision.

\subsection{Training Scheme}
\label{sec:training}

First, we generate training and validation dataset as in \cite{bevlocator}. We add random noise to real vehicle poses with longitudinal deviation $\triangle x^*$, lateral deviation $\triangle y^*$, and yaw deviation $\triangle \alpha^*$. The HD maps $\mathcal{M}$ are transformed and cropped by the noised vehicle poses, and fed into the proposed network as $\mathcal{M}_0$. 

For stable and fast convergence in network training, we conduct a two-stage training scheme. In the first stage, we train the BEV perception backbone, the rasterized map encoder, and the \textbf{\texttt{FLORA}} module through the semantic supervision $\mathcal{L}_{sem}$ \cite{bemapnet} on both $\hat{\textbf{M}}^e_\mathcal{I}$ and $\hat{\textbf{M}}_\mathcal{M}$.
This stage ensures that the enhanced BEV features and the rasterized map features contain implicit semantic map element information, which is matchable for the pose solver.
We do not add additional supervision on the estimated flow from the \textbf{\texttt{FLORA}} module and only guide it to enhance the BEV features through self-learning flow estimation.

Then, in the second stage, we further add the probability supervision $\mathcal{L}_{prob}$ on the coarse \textbf{\texttt{DEMA}} module and the pose supervision $\mathcal{L}_{pose}$ on the fine \textbf{\texttt{POET}} module for E2E network training.
The localization network aims to estimate the added pose deviations ($\triangle \textbf{T}^{*}=[\triangle x^*, \triangle y^*, \triangle \alpha^*]$) based on $\mathcal{M}_0$ and $\mathcal{I}$. 
Therefore, $\triangle \textbf{T}^*$ can be utilized as the supervision for pose estimation, allowing the network to be E2E trained.

Concretely, the proposed \textbf{\texttt{DEMA}} module provides the probability distribution of 3 DoF pose hypotheses, \textit{i.e.}, $\hat{\textbf{P}}_x=\{p(\triangle {x}_i)\}^{N_x}_{i=1},\hat{\textbf{P}}_y=\{p(\triangle {y}_j)\}^{N_y}_{j=1},\hat{\textbf{P}}_\alpha=\{p(\triangle \alpha_k)\}^{N_\alpha}_{k=1}$. Generally, we hope the pose hypotheses near the ground-truth pose $\triangle \textbf{T}^*$ have higher probabilities while decreasing the probabilities of other pose hypotheses. Thus, we add probability supervisions $\mathcal{L}_{prob}$ on the probability distributions of 3 DoF pose hypotheses. For example, we sample the probability of ground-truth yaw deviation $\triangle \alpha^*$ from the probability distribution of yaw hypotheses $\hat{\textbf{P}}_\alpha$, and calculate its negative logarithmic value as the probability loss:
\begin{equation}
    \mathcal{L}_{prob}(\hat{\textbf{P}}_\alpha)=-\texttt{log}(\texttt{sample1d}(\hat{\textbf{P}}_\alpha,\triangle \alpha^*))
\end{equation}
As the loss decreases, the probability of $\triangle \alpha^*$ will increased so that the probability distributions from the \textbf{\texttt{DEMA}} module will gradually turn into peaky distributions centered on the ground-truth poses as training proceeds. 
The probability losses for longitudinal and lateral positions are similar to the yaw angle. 

As for the fine pose estimation from the proposed \textbf{\texttt{POET}} module, we calculate the smooth L1 distance between $\triangle \textbf{T}_f$ and $\triangle \textbf{T}^*$ as the pose loss:
\begin{equation}
    \mathcal{L}_{pose}(\triangle \textbf{T}_f)={||\triangle \hat{\textbf{T}}_f-\triangle \textbf{T}^*||}_{S1}
\end{equation}

Thus, the final loss in the second E2E training stage is:
\begin{align}
    \mathcal{L} =&\mathcal{L}_{sem}(\hat{\textbf{M}}^e_\mathcal{I})+\mathcal{L}_{sem}(\hat{\textbf{M}}_\mathcal{M}) +w\cdot \mathcal{L}_{pose}(\triangle \hat{\textbf{T}}_f) \notag \\
     &+0.1\cdot (\mathcal{L}_{prob}(\hat{\textbf{P}}_x) + \mathcal{L}_{prob}(\hat{\textbf{P}}_y) + \mathcal{L}_{prob}(\hat{\textbf{P}}_\alpha))
\end{align}

where $w$ is an adaptive weight factor based on relative value $r=\mathcal{L}_{pose}(\triangle \textbf{T}_f) / \mathcal{L}_{sem}(\hat{\textbf{M}}^e_\mathcal{I})$:
\begin{equation}
w = \left\{\begin{array}{rcl}
        10 & \mbox{if} & r < 0.1 \\
        0.1 & \mbox{if} & r > 10 \\
        1 & & \mbox{otherwise}
    \end{array}\right.
\end{equation}

\subsection{Adaptive Coarse-to-fine Localization}
\label{sec:adaptive}

Although refining the coarse pose by the Transformer-based \textbf{\texttt{POET}} module can effectively improve localization accuracy, it inevitably impose significantly more computational burden than utilizing the \textbf{\texttt{DEMA}} module only.
To achieve a balance between localization performance and efficiency, we leverage the estimated probability of the \textbf{\texttt{DEMA}} module as an estimated reliability of the coarse pose, adaptively triggering the \textbf{\texttt{POET}} module when needed.

Concretely, if the estimated probability (reliability) of the coarse pose is greater than the predefined threshold $t_c$, we believe the coarse pose is sufficiently reliable and do not refine it through the fine localization module:
\begin{equation}
\triangle \hat{\textbf{T}} = \left\{\begin{array}{rcl}
        \triangle \hat{\textbf{T}}_c & \mbox{if} & \hat{\textbf{P}}_c \ge t_c \\
        \triangle \hat{\textbf{T}}_f & & \mbox{otherwise}
    \end{array}\right.
\end{equation}
By applying such an adaptive strategy, the heavy \textbf{\texttt{POET}} module is not activated when the \textbf{\texttt{DEMA}} module already estimates vehicle pose reliably, reducing unnecessary computational overhead in the E2E localization pipeline.

%% file: secs/4-experiment.tex
\section{Experiments}

\subsection{Implementation Details}

{
\setlength{\parindent}{0cm}
\textbf{Model:}
We implemented the proposed network using the PyTorch library. 
The proposed network is trained using AdamW optimizer, a batch size of 8, an initial learning rate of $1e{-4}$, a weight decay rate of $1e{-7}$ on 4 NVIDIA RTX 3090 GPUs. 
The first stage trains for 10 epochs, followed by 15 epochs for full model training in the second stage.
Following BeMapNet \cite{bemapnet}, the BEV perception focus on three kinds of static map elements, \textit{i.e.}, lane dividers, pedestrian crossings, and road boundaries. 
The perception ranges covers $\left[-15.0m, 15.0m\right]$ laterally and $\left[-30.0m, 30.0m\right]$ longitudinally.
The resolution of BEV grid is set as 0.15m/pixel, resulting in $H\times W=400 \times 200$ HD map masks.
For fair comparison, the sampling ranges of pose hypotheses are set to $\pm 2m,~\pm1m,~\pm2\degree$ in longitudinal, lateral, and yaw dimension as \cite{bevlocator}, and the sampling steps are set to $0.4m,~0.2m$ and $0.4\degree$ for efficiency if not specific. Therefore, the amounts of pose hypotheses in each dimension all equal to 11.
The feature dimensions and sizes are set as $H_f\times W_f=50\times 25,D=128,H_c\times W_c=100\times 50,D_c=16$, and $K_f=16$ by default.
We utilize 6 Transformer encoder layers and 6 Transformer decoder layers in the proposed \textbf{\texttt{POET}} module.
The full model with adaptive strategy can run in approximately 150 ms on a commercial NVIDIA RTX 3090 GPU.
}

{
\setlength{\parindent}{0cm}
\textbf{Benchmark:}
The nuScenes dataset \cite{caesar2020nuscenes} is utilized in this paper, which consist of 28,130 training and 6,019 validation samples from 700 training and 150 validation driving scenes, respectively.
The dataset covers different regions (urban, residential, nature, and industrial), times (day and night), and weather conditions (sunny, rainy, and cloudy), presenting a significant challenge for evaluating generalization capability of the perception and localization methods \cite{hdmapnet,bemapnet}.
Only 6 surrounding cameras are used in the visual localization task.
During the training and validation process, we add random pose deviations to real vehicle poses as the initial vehicle pose. 
The ranges of randomly generated longitudinal, lateral, and yaw deviations are set to $\pm 2m, \pm 1m, \pm 2\degree$. We measure the localization accuracy of estimated 3 DoF poses \cite{bevlocator}, and Intersection-over-Union (IoU) scores of perceived static map elements \cite{hdmapnet} for quantitative evaluation and comparison.
}

\begin{table}[!t]
\centering
\caption{Comparison of localization accuracy with existing methods in terms of longitudinal, lateral and yaw errors.}
\renewcommand\arraystretch{1.5}
\begin{threeparttable}
\resizebox{\linewidth}{!}{
\begin{tabular}{l c c c c c}
\toprule
\multirow{2}{*}{Methods} & \multirow{2}{*}{Sensors} & \multirow{2}{*}{Dataset} & \multicolumn{3}{c}{Localization error $\downarrow$} \\
& & & $x~(m)$ & $y~(m)$ & $\alpha~(\degree)$ \\
\midrule
\multicolumn{6}{l}{\textbf{Measured by Root Mean Square Error (RMSE)}} \\
\hline
\multirow{2}{*}{Pauls et al. \cite{pauls2020monocular}} & \multirow{2}{*}{mono.} & {Karlsruhe} & 0.86 & 0.23 & 0.58 \\
& & {Ludwigsburg} & 0.90 & 0.11 & 0.56 \\
\multirow{2}{*}{Xiao et al. \cite{xiao2018monocular}} & \multirow{2}{*}{mono.} & scenario 1 & \multicolumn{2}{c}{0.27} & - \\
 & & scenario 2 & \multicolumn{2}{c}{0.37} & - \\
\multirow{2}{*}{Xiao et al. \cite{xiao2020monocular}} & \multirow{2}{*}{mono.} & scenario 1 & \multicolumn{2}{c}{0.21} & - \\
 & & scenario 2 & \multicolumn{2}{c}{0.29} & - \\
Zhao \textit{et al.} \cite{Zhao2024Towards} & surro. & nuScenes & \multicolumn{2}{c}{0.34} & - \\
ICP \cite{Zhao2024Towards} & surro. & nuScenes & \multicolumn{2}{c}{0.39} & - \\
Miao \textit{et al.} \cite{miao2025efficient} & surro. & nuScenes & 0.37 & 0.21 & 0.59 \\ 

\rowcolor{gray!20} Ours (C) & surro. & nuScenes & 0.244 & 0.189 & 0.538 \\
\rowcolor{gray!20} Ours (F) & surro. & nuScenes & 0.290 & 0.222 & 0.520 \\
\rowcolor{gray!20} Ours (C+F) & surro. & nuScenes & 0.225 & 0.187 & 0.526 \\
\rowcolor{gray!20} Ours (A) & surro. & nuScenes & 0.229 & 0.187 & 0.531 \\

\hline
\multicolumn{6}{l}{\textbf{Measured by Mean Absolute Error (MAE)}} \\
\hline
\multirow{2}{*}{Pauls et al. \cite{pauls2020monocular}} & \multirow{2}{*}{mono.} & {Karlsruhe} & 0.70 & 0.19 & 0.45 \\
& & {Ludwigsburg} & 0.73 & 0.07 & 0.28 \\
Wang et al. \cite{wang2021visual} & mono.  & urban & 0.43 & 0.12 & 0.11 \\
OrienterNet-SD \cite{sarlin2023orienternet} & mono. & nuScenes & \multicolumn{2}{c}{14.79} & 46.04 \\
SegLocNet-SD \cite{zhou2025seglocnet} & surro. & nuScenes & \multicolumn{2}{c}{8.15} & 19.68 \\
SegLocNet-HD \cite{zhou2025seglocnet} & surro. & nuScenes & \multicolumn{2}{c}{5.30} & 10.11 \\
\multirow{2}{*}{Zhang et al. \cite{zhang2022learning}} & \multirow{2}{*}{mono.} & Highway & 0.20 & 0.05 & - \\
& & nuScenes & 0.22 & 0.10 & - \\
BEVLocator \cite{bevlocator} & surro. & nuScenes & 0.18 & 0.08 & 0.51 \\
Miao \textit{et al.} \cite{miao2025efficient} & {surro.} & {nuScenes} & 0.19 & 0.13 & 0.39  \\ 
RTMap (E2E) \cite{du2025rtmap} & {surro.} & {nuScenes} & 0.589 & 0.142 & 0.521 \\
RTMap (opt) \cite{du2025rtmap} & {surro.} & {nuScenes} & 0.586 & 0.121 & 0.368 \\

\rowcolor{gray!20} Ours (C) & surro. & nuScenes & 0.145 & 0.129 & 0.394 \\
\rowcolor{gray!20} Ours (F) & surro. & nuScenes & 0.171 & 0.165 & 0.389 \\
\rowcolor{gray!20} Ours (C+F) & surro. & nuScenes & 0.125 & 0.128 & 0.387 \\
\rowcolor{gray!20} Ours (A) & surro. & nuScenes & 0.132 & 0.128 & 0.392 \\
\bottomrule
\end{tabular}}
\begin{tablenotes}
    \footnotesize
    \item[] ``mono.'' is monocular camera while ``surro.'' is 6 surrounding cameras.
    \item[] ``C'' uses the coarse localization module only.
    \item[] ``C+F'' uses full coarse-to-fine localization system.
    \item[] ``A'' uses adaptive coarse-to-fine localization system.
\end{tablenotes}
\end{threeparttable}
\label{table:main}
\vspace{-11pt}
\end{table}

\subsection{Comparison with the State-of-the-Arts}

In order to prove the effectiveness and novelty of the proposed E2E localization network \textbf{\texttt{RAVE}}, we evaluate the localization accuracy in terms of the estimated 3 DoF poses on the nuScenes dataset \cite{caesar2020nuscenes} and compare our method with existing rule-based visual localization methods \cite{pauls2020monocular,xiao2018monocular,xiao2020monocular,wang2021visual,Zhao2024Towards} and learning-based methods \cite{bevlocator,miao2025efficient,sarlin2023orienternet,zhou2025seglocnet,du2025rtmap}. For fairness, we report the used datasets and evaluation metrics of these compared methods, since some methods are tested on self-collected data and not open-sourced, and they are evaluated using different evaluation metrics, such as Root Mean Square Error (RMSE) and Mean Absolute Error (MAE). 
We also test the performance of the E2E localization network with different configurations, that is, the one using coarse localization \textbf{\texttt{DEMA}} only (\texttt{Ours(C)}), the one using fine localization \textbf{\texttt{POET}} only (\texttt{Ours(F)}), the full system without adaptive strategy (\texttt{Ours(C+F)}), and the adaptive localization system (\texttt{Ours(A)} or \textbf{\texttt{RAVE}}), as listed in Tab. \ref{table:main}.

\begin{figure*}[!t]
    \centering
    \includegraphics[width=0.97\linewidth]{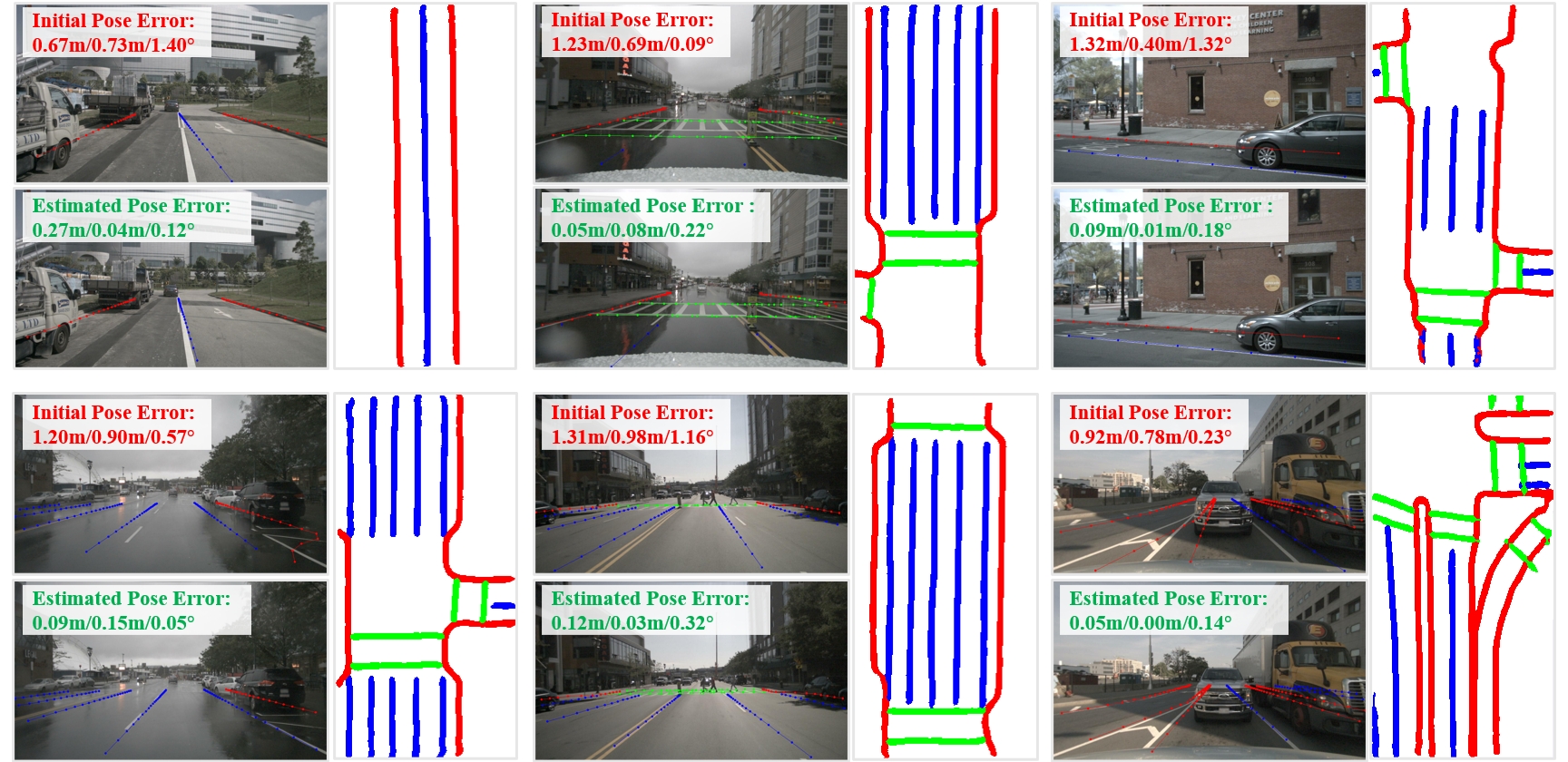}
    \caption{Visualization cases of E2E localization on the nuScenes \cite{caesar2020nuscenes}. HD map elements are projected onto images viz initial pose (upper left) and estimated pose (bottom left). The perceived map elements are shown in BEV space (right). Road boundaries, lane dividers, and pedestrian crossings are drawn in red, blue, and green, respectively.}
    \label{fig:main}
\end{figure*}

The quantitative experimental results show that our proposed method \texttt{\textbf{RAVE}} can achieve decimeter-level accuracy with MAE of 0.132m in longitudinal position, 0.128m in lateral position, and 0.392$\degree$ in yaw angle on the nuScenes dataset \cite{caesar2020nuscenes}. 
Compared to those rule-based localization methods \cite{pauls2020monocular,xiao2018monocular,xiao2020monocular,wang2021visual}, our method achieves comparable or even improved localization accuracy. It should be noticed that our method is evaluated on a more challenging benchmark, where weather and lighting conditions are significantly varied, compared with the simple and single evaluation scenarios utilized in rule-based methods. 
Compared to the method proposed by Zhao \textit{et al.} \cite{Zhao2024Towards} that also utilizes learning-based BEV perception, our work localizes vehicles in an E2E scheme through deep neural networks including both BEV perception and differentiable pose estimation, so the parameters of pose estimation network can be automatically tuned throughout the E2E training stage, leading to better robustness and localization accuracy. 
This indicates the advantages and potentials of the full E2E localization methods that after training on large volumes of real-world data, the network can generalize well on varying realistic driving scenarios without scene-specific parameter and rule tuning by experienced engineers.

Compared to learning-based E2E localization algorithms \cite{sarlin2023orienternet,zhou2025seglocnet,bevlocator,miao2025efficient,du2025rtmap}, our work also demonstrates significantly improved localization accuracy.
OrienterNet \cite{sarlin2023orienternet} and SegLocNet \cite{zhou2025seglocnet} utilize \textbf{\texttt{FUMA}} solver to estimate pose, but their localization accuracy is limited due to the utilized SD map and the large initial pose error. Our work improves the efficiency of neural matching by the proposed \textbf{\texttt{DEMA}} solver, which can be easily plugged into these works.
While Miao \textit{et al.} utilized \textbf{\texttt{DEMA}} solver for efficient pose estimation \cite{miao2025efficient}, our work in this paper substantially extends it by novel map prior-enhanced perception and hierarchical localization modules that improves the robustness of perception and the accuracy of localization, especially in terms of longitudinal pose estimation.
BEVLocator \cite{bevlocator} and RTMap \cite{du2025rtmap} utilize full Transformer-based pose regressor and achieve superior lateral pose estimation accuracy to ours, but their performance on longitudinal position and yaw angle estimation is limited. Moreover, the whole black-box framework in BEVLocator \cite{bevlocator} and RTMap \cite{du2025rtmap} lacks reliability and interpretability, which brings uncertainty to AD tasks. Our proposed \textbf{\texttt{RAVE}} algorithm first estimates coarse vehicle pose by interpretable \textbf{\texttt{DEMA}} solver and then refines the estimation by Transformer-based \textbf{\texttt{POET}} module, which achieves localization accuracy, efficiency, and interpretability.

We provide some qualitative visualization cases in Fig. \ref{fig:main}. Our method can jointly perceive the static driving environment and perform localization under varying environmental conditions through a unified E2E network. By applying our method, vehicle pose can be accurately estimated, enabling precise alignment between projected map elements and realistic landmarks, even with large initial pose errors. 

\begin{table}[!t]
\centering
\caption{Robustness towards various visual conditions. The improved performance by fine localization is highlighted by \textbf{BOLD}.}
\renewcommand\arraystretch{1.5}
\begin{threeparttable}
\resizebox{\linewidth}{!}{
\begin{tabular}{l c c c c c c}
\toprule
\multirow{2}{*}{MAE $\downarrow$} & \multicolumn{6}{c}{Environmental Condition} \\
& day & night & sunny & cloudy & rainy & all \\
\midrule
\multicolumn{7}{l}{\texttt{Ours (C)}} \\
\hline
x (m) & 0.147 & 0.135 & 0.142 & 0.145 & 0.164 & 0.145  \\
y (m) & 0.128 & 0.134 & 0.130 & 0.121 & 0.133 & 0.129  \\
$\alpha~(\degree)$ & 0.394 & 0.390 & 0.413 & 0.358 & 0.394 & 0.394\\
\hline
\multicolumn{7}{l}{\texttt{Ours (A)}} \\
\hline
x (m) & \textbf{0.133} & \textbf{0.122} & \textbf{0.128} & \textbf{0.133} & \textbf{0.149} & \textbf{0.132}  \\
y (m) & \textbf{0.127} & \textbf{0.133} & \textbf{0.129} & 0.121 & \textbf{0.132} & \textbf{0.128}  \\
$\alpha~(\degree)$ & \textbf{0.392} & \textbf{0.386} & \textbf{0.410} & \textbf{0.355} & 0.399 & \textbf{0.392} \\
\bottomrule
\end{tabular}}
\end{threeparttable}
\label{table:weather_loc}
\end{table}

We also test the robustness of the proposed method towards various environmental conditions, as shown in Tab. \ref{table:weather_loc}. Taking advantage of the E2E framework, the network achieves stable pose estimation with high precision under varying conditions. In most cases, the fine localization stage in the proposed hierarchical localization framework can refine the estimated pose and improve the localization accuracy.

\subsection{Ablation Analysis}

\subsubsection{Ablation Analysis about the \texttt{\textbf{FLORA}} module}
To analyze the proposed map prior-based BEV perception enhancement method, \textbf{\texttt{FLORA}}, we evaluate its performance and determine its optimal configuration through quantitative experiments. 

\begin{table}[!t]
\centering
\caption{Comparison of online BEV perception performance with existing methods.}
\renewcommand\arraystretch{1.5}
\begin{threeparttable}
\resizebox{\linewidth}{!}{
\begin{tabular}{l c c c c c}
\toprule
\multirow{2}{*}{Methods} & \multicolumn{5}{c}{BEV Perception performance (IoU) $\uparrow$} \\
& $\mbox{bg}.~(\%)$ & $\mbox{div.}~(\%)$ & $\mbox{ped.}~(\%)$ & $\mbox{boud.}~(\%)$ & $\mbox{avg.}~(\%)$ \\
\midrule
\multicolumn{6}{l}{\textbf{Vision-only solution}} \\
\hline
HDMapNet \cite{hdmapnet} & - & 40.6 & 18.7 & 39.5 & - \\
BeMapNet \cite{bemapnet} & 88.5 & 46.5 & 27 & 46 & 52 \\
\hline
\multicolumn{6}{l}{\textbf{Map Prior Enhanced solution}} \\
\hline
P-MapNet (+S+H) \cite{jiang2024pmapnet} & - & 44.3 & 23.3 & 43.8 & - \\
PriorDrive (+H) \cite{zeng2024driving} & - & 42.3 & 21.2 & 44.9 & -\\
NMP (+N) \cite{nmp} & - & 55.0 & 34.1 & 56.5 & - \\
DiffMap (+H) \cite{jia2024diffmap} & - & 42.1 & 23.9 & 42.2 & - \\
\rowcolor{gray!10} BeMapNet-ADD (+H) & 90.5 & 57.9 & 27.5 & 52.7 & 57.1 \\
\rowcolor{gray!10} BeMapNet-CON (+H) & 90.3 & 56.9 & 26.4 & 51.8 & 56.4 \\
\rowcolor{gray!10} BeMapNet-ATT (+H) & 88 & 46.8 & 34.5 & 48.3 & 54.4 \\
\rowcolor{gray!10} BeMapNet-GRU (+H) & 90.4 & 57.7 & 28.4 & 53.8 & 57.6 \\
\rowcolor{gray!10} BeMapNet-DGRU (+H) & 88.1 & 47 & 57 & 48 & 60 \\
\rowcolor{gray!10} BeMapNet-FLOW (+H) & 93.1 & 65.8 & 65.2 & 66.1 & 72.5 \\
\hline
\multicolumn{6}{l}{\textbf{Ours (+H)} } \\
\hline
\rowcolor{gray!20} BeMapNet-FLORA (4) & 91.5 & 59.3 & 63.6 & 60.1 & 68.6 \\
\rowcolor{gray!20} BeMapNet-FLORA (8) & 91.8 & 60.1 & 64.8 & 61.1 & 69.4 \\
\rowcolor{gray!20} BeMapNet-FLORA (12) & 92.3 & 62.5 & 66.2 & 62.8 & 71 \\
\rowcolor{gray!40} BeMapNet-FLORA (16) & 93.7 & 67.9 & 68.6 & 68.4 & 74.6 \\
\rowcolor{gray!20} BeMapNet-FLORA (20) & 91.3 & 58.3 & 63.9 & 59.5 & 68.2 \\
\rowcolor{gray!20} BeMapNet-FLORA (24) & 89.4 & 53.8 & 41.7 & 51 & 59 \\

\bottomrule
\end{tabular}}
\begin{tablenotes}
    \footnotesize
    \item[] ``S'' is SD map prior, ``H'' is HD map prior, and ``N'' is neural map prior.
\end{tablenotes}
\end{threeparttable}
\label{table:bev}
\end{table}

\begin{table}[!t]
\centering
\caption{The improved robustness towards various visual conditions using the \textbf{\texttt{FLORA}} module.}
\renewcommand\arraystretch{1.5}
\begin{threeparttable}
\resizebox{\linewidth}{!}{
\begin{tabular}{l c c c c c}
\toprule
\multirow{2}{*}{Methods} & \multicolumn{5}{c}{BEV Perception performance (IoU) $\uparrow$} \\
& $\mbox{bg}.~(\%)$ & $\mbox{div.}~(\%)$ & $\mbox{ped.}~(\%)$ & $\mbox{boud.}~(\%)$ & $\mbox{avg.}~(\%)$ \\
\midrule
\multicolumn{6}{l}{BeMapNet \cite{bemapnet}} \\
\hline
day & 88.1 & 46.5 & 27.3 & 46 & 52  \\
night & 91.7 & 46.9 & 12.3 & 45.2 & 49  \\
sunny & 87.9 & 45.3 & 30.1 & 46.9 & 52.6\\
cloudy & 88.8 & 48.6 & 24.2 & 47.9 & 52.4  \\
rainy & 87.4 & 46.7 & 22.9 & 39.7 & 49.2 \\
\hline
\multicolumn{6}{l}{BeMapNet-FLORA (16)} \\
\hline
day & 93.5 & 67.9 & 68.6 & 68.4 & 74.6  \\
night & 95.3 & 67.9 & 70.6 & 68.5 & 75.6  \\
sunny & 93.3 & 67.2 & 69.2 & 68.3 & 74.5\\
cloudy & 93.8 & 69.4 & 68 & 69 & 75  \\
rainy & 93.3 & 67.7 & 67.3 & 67.4 & 73.9 \\

\bottomrule
\end{tabular}}
\end{threeparttable}
\label{table:weather}
\end{table}

\paragraph{Effectiveness of the \textbf{\texttt{FLORA}} module}
In this paper, BeMapNet \cite{bemapnet} is selected as our BEV perception backbone. We compare this baseline with some map prior-based perception enhancement works \cite{jiang2024pmapnet,zeng2024driving,nmp,jia2024diffmap} and our proposed \textbf{\texttt{FLORA}} method. Since BEV perception enhanced by rasterized HD map lacks existing methods, we also build some compared methods using well-known fusion techniques:
\begin{itemize}
    \item BeMapNet-ADD: $\textbf{F}^e_\mathcal{I}=\textbf{F}_\mathcal{I}+\textbf{F}^R_\mathcal{M}$;
    \item BeMapNet-CON: $\textbf{F}^e_\mathcal{I}=\texttt{conv}([\textbf{F}_\mathcal{I},\textbf{F}^R_\mathcal{M}])$;
    \item BeMapNet-ATT: $\textbf{F}^e_\mathcal{I}=\texttt{crossattn}(\textbf{F}_\mathcal{I},\textbf{F}^R_\mathcal{M})$;
    \item BeMapNet-GRU: $\textbf{F}^e_\mathcal{I}=\texttt{GRU}(\textbf{F}_\mathcal{I},\textbf{F}^R_\mathcal{M})$;
    \item BeMapNet-DGRU: $\textbf{F}^e_\mathcal{I}=\texttt{DeformableGRU}(\textbf{F}_\mathcal{I},\textbf{F}^R_\mathcal{M})$;
    \item BeMapNet-FLOW acts as \texttt{\textbf{FLORA}}, but it does not perform low-rank flow feature projection.
\end{itemize}
The experimental results are listed in Tab. \ref{table:bev}.

From the table, we can see that the BEV perception performance is consistently improved by the map prior. P-MapNet \cite{jiang2024pmapnet} and DiffMap \cite{jia2024diffmap} learn HD map prior in their training stage, resulting in limited perception enhancement. PriorDrive \cite{zeng2024driving} fuses the HD map prior online, but uses only the road boundaries. In localization tasks with available HD maps, these methods cannot fully leverage the HD map prior to enhance online BEV perception. However, without precise vehicle pose, the retrieved HD map is not accurately aligned to online perception. Directly fusing HD map prior leads to unpleasant performance, as shown by BeMapNet-ADD, BeMapNet-CON, BeMapNet-ATT, and BeMapNet-GRU in Tab. \ref{table:bev}. If the spatial displacement between retrieved HD map and online perception is solved, the perception enhancement will be significant, like BeMapNet-DGRU and BeMapNet-FLOW.

Experimental results demonstrate the proposed \textbf{\texttt{FLORA}} module achieves the best performance in BEV perception. With the low-rank flow feature projection, the dominant flow feature can be retained, while noise will be reduced, enabling more accurate alignment between HD map prior and online perception, leading to better enhancement than vanilla flow. The experiment shows the \textbf{\texttt{FLORA}} module improves perception performance in terms of IoU scores from 52\% to 74.6\%.

\begin{figure}[!t]
    \centering
    \includegraphics[width=0.97\linewidth]{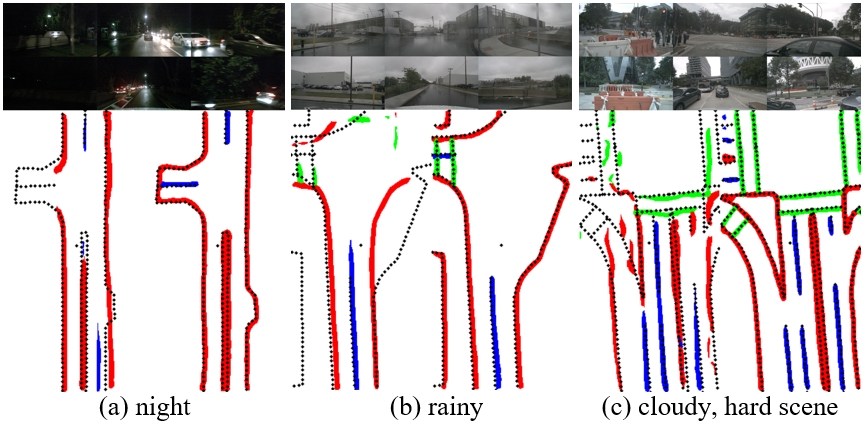}
    \caption{Visualization cases of the enhanced BEV perception on the nuScenes \cite{caesar2020nuscenes}. The upper row shows surrounding images. The perceived map elements by original BeMapNet \cite{bemapnet} (bottom row: left) and BeMapNet-FLORA (bottom row: right) are shown in BEV space. Road boundaries, lane dividers, and pedestrian crossings are drawn in red, blue, and green, respectively. Ground-truth HD map points are visualized in BEV space using black circles.}
    \label{fig:bev}
\end{figure}

We also evaluate the perception enhancement by \textbf{\texttt{FLORA}} module under varying environmental conditions, as shown in Tab. \ref{table:weather} and Fig. \ref{fig:bev}. The perception performance of original BeMapNet \cite{bemapnet} in terms of difficult map elements (such as pedestrian crossing) decreases dramatically under poor lighting (night), cloudy, and rainy conditions. With \textbf{\texttt{FLORA}} module, the HD map prior effectively enhances online perception so that the system can perceive stably and accurately under different conditions with a consistent improvement of more than 20\% in terms of IoU scores.

\paragraph{Effectiveness of Flow Feature Rank}
In the proposed \textbf{\texttt{FLORA}} module, the high-dimensional flow features are projected into the subspace built by low-rank flow features bases to suppress noise in flow estimation. To determine the optimal configuration of the \textbf{\texttt{FLORA}} module, we regulate the rank of flow feature, $K$, and test the perception performance of the variants, as in Tab. \ref{table:bev}. According to the results, as flow feature rank increases, the perception performance initially improves and then declines, showing that a proper rank of flow feature is necessary for flow estimation. Based on the experimental results, the \textbf{\texttt{FLORA}} module with $K=16$ achieves the best perception performance, which is chosen as the default configuration in this paper.

\subsubsection{Ablation Analysis about the \textbf{\texttt{DEMA}} module}
\begin{figure}[!t]
    \centering
    \includegraphics[width=0.97\linewidth]{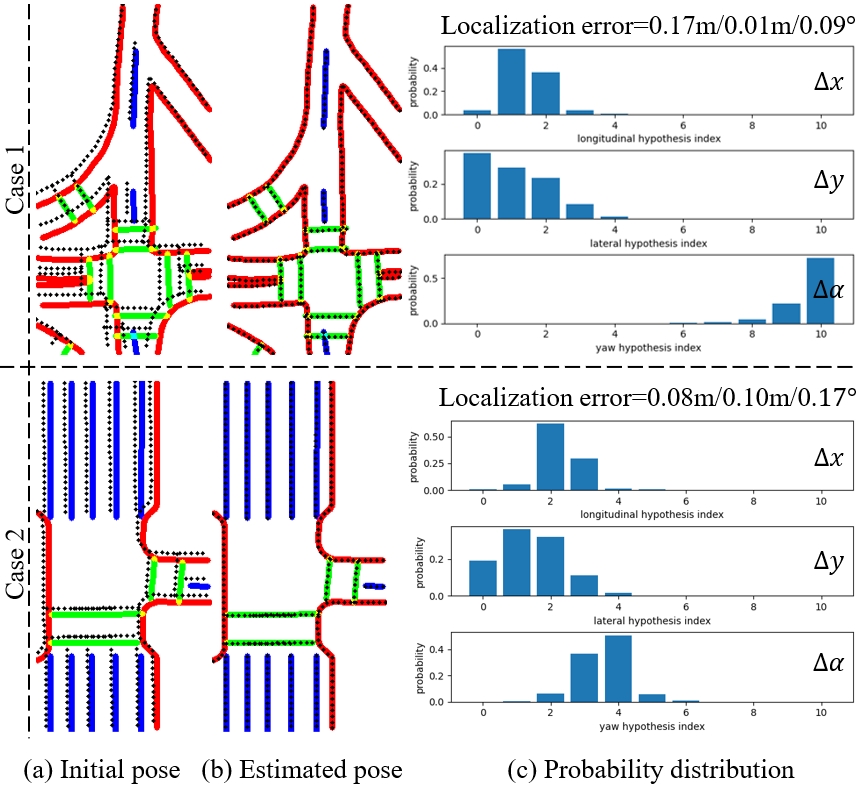}
    \caption{Some examples of the \textbf{\texttt{DEMA}} solver. (a) illustrates the map projection using initial pose, (b) displays the map projection using estimated pose, (c) presents the estimated probability distribution of the 3 DoF poses. In (a) and (b),  projected HD map points are drawn in black circles. Ground-truth HD map masks are drawn in blue, green, and red lines, corresponding to lane dividers, pedestrian crossings, and road boundaries, respectively.}
    \label{fig:coarse}
\end{figure}

To analyze the proposed \textbf{\texttt{DEMA}} solver and conclude its novelty, we compare its performance with traditional \textbf{\texttt{FUMA}} solver and then evaluate the performance of variants with different inputs, different sizes of input features, and different sampling steps. To intuitively visualize the estimated probabilities of 3 DoF poses by \textbf{\texttt{DEMA}} solver, we provide an example in Fig. \ref{fig:coarse}. Following Sec. \ref{sec:localization}, the decoupled pose hypothesis with the highest probability is selected as the final estimation of \textbf{\texttt{DEMA}} solver.
For rapid evaluation, the \textbf{\texttt{FUMA}} solver and \texttt{\textbf{DEMA}} solver evaluated in this part are trained for only 5 epochs in the second E2E training stage without adding the \textbf{\texttt{POET}} module.

\begin{table*}[!t]
	\centering
	\caption{The localization performance using various pose solver and different inputs.}
    \renewcommand\arraystretch{1.5}
	\begin{threeparttable}
    \resizebox{\linewidth}{!}{
		\begin{tabular}{l | l | c | c |c | c | c | c c c | r }
			\toprule
			& \multirow{2}{*}{Solver} & \multirow{2}{*}{\textbf{\texttt{FLORA}}} & \multirow{2}{*}{Inputs} & \multirow{2}{*}{$D_c$} & 
            \multirow{2}{*}{BEV Size} & 
            {Sampling Step} & 
            \multicolumn{3}{c | }{Localization MAE $\downarrow$ } & Memory${}^1$ \\
            & & & & & & (m / m / \degree) & $x~(cm)$ & $y~(cm)$ & $\alpha~(\degree)$ & (MiB) \\
			\midrule
            (F.a) & FUMA &  & $\textbf{F}_\mathcal{I},\textbf{F}^R_\mathcal{M}$ & 16 & $100\times50$ & 0.4/0.2/0.4 & 17.3 & 16.4 & 0.411 & 64.31  \\ 
            (F.b) & FUMA${}^2$ &  & $\textbf{F}_\mathcal{I},\textbf{F}^R_\mathcal{M}$ & 16 & $100\times50$ & 0.2/0.1/0.2 & 14.5 & 15.7 & 0.397 & 14888.95 \\ 
            (F.c) & FUMA &  & $\textbf{F}_\mathcal{I},\textbf{F}^R_\mathcal{M}$ & 16 & $100\times50$ & 0.8/0.4/0.8 & 25.2 & 18.2 & 0.453 & 8.73 \\ 
            (F.d) & FUMA & \checkmark & $\textbf{F}^e_\mathcal{I},\textbf{F}^R_\mathcal{M}$ & 16 & $100\times50$ & 0.4/0.2/0.4 & 15.9 & 14.9 & 0.427 & 66.15  \\ 
            \hline
            (D.a) & DEMA & & $\hat{\textbf{M}}_\mathcal{I},\hat{\textbf{M}}^R_\mathcal{M}$ & 4 & $100\times50$ & 0.4/0.2/0.4 & 23.1 & 18.6 & 0.533 & 1.82  \\ 
            (D.b) & DEMA & \checkmark & $\hat{\textbf{M}}^e_\mathcal{I},\hat{\textbf{M}}^R_\mathcal{M}$ & 4 & $100\times50$ & 0.4/0.2/0.4 & 16.7 & 14.8 & 0.496 & 1.85  \\ 
            (D.c) & DEMA & & $\textbf{F}_\mathcal{I},\textbf{F}^R_\mathcal{M}$ & 16 & $100\times50$ & 0.4/0.2/0.4 & 16.1 & 16.7 & 0.495 &  2.53 \\ 
            (D.d) & DEMA & & $\textbf{F}_\mathcal{I},\textbf{F}^R_\mathcal{M}$ & 16 & $100\times50$ & 0.2/0.1/0.2 & 13.1 & 16.3 & 0.502 &  3.67 \\ 
            (D.e) & DEMA & & $\textbf{F}_\mathcal{I},\textbf{F}^R_\mathcal{M}$ & 16 & $100\times50$ & 0.8/0.4/0.8 & 24.9 & 19.4 & 0.517 &  1.96 \\ 
            (D.f) & DEMA & & $\textbf{F}_\mathcal{I},\textbf{F}^R_\mathcal{M}$ & 16 & $400\times200$ & 0.4/0.2/0.4 & 16.2 & 15.8 & 0.462 & 22.84 \\ 
            \rowcolor{gray!20}(D.g) & DEMA & \checkmark & $\textbf{F}^e_\mathcal{I},\textbf{F}^R_\mathcal{M}$ & 16 & $100\times50$ & 0.4/0.2/0.4 & 16.0 & 14.2 & 0.424 & 2.14 \\ 
            (D.h) & DEMA & & $\textbf{F}_\mathcal{I},\textbf{F}^R_\mathcal{M}$ & 64 & $100\times50$ & 0.4/0.2/0.4 & 15.5 & 15.4 & 0.490 & 10.69 \\ 
            (D.i) & DEMA & \checkmark & $\textbf{F}^e_\mathcal{I},\textbf{F}^R_\mathcal{M}$ & 64 & $100\times50$ & 0.4/0.2/0.4 & 15.0 & 13.6 & 0.419 & 11.00 \\ 
            (D.j) & DEMA & & $\textbf{F}_\mathcal{I},\textbf{F}^R_\mathcal{M}$ & 128 & $100\times50$ & 0.4/0.2/0.4 & 15.2 & 14.6 & 0.471 & 36.15  \\ 
            (D.k) & DEMA & \checkmark & $\textbf{F}^e_\mathcal{I},\textbf{F}^R_\mathcal{M}$ & 128 & $100\times50$ & 0.4/0.2/0.4 & 15.1 & 13.6 & 0.422 & 35.36 \\ 
			\bottomrule
		\end{tabular}
    }
	\end{threeparttable}
    \begin{tablenotes}
        \footnotesize
        \item ${}^1$We measure the inference memory required by the the localization module by subtracting the GRU memory of BEV perception from that of the entire network.   
        \item ${}^2$The solvers are evaluated with small sampling step using the trained model (F.a) due to unacceptable memories required by training process.
    \end{tablenotes}
	\label{table:coarse}
\end{table*}

\paragraph{\textbf{\texttt{FUMA}} v.s. \textbf{\texttt{DEMA}}}
First, we compare the localization performance and computational efficiency between traditional \textbf{\texttt{FUMA}} solver and the proposed \textbf{\texttt{DEMA}} solver. The results are listed as Tab. \ref{table:coarse} (F.a)-(D.c) and (F.d)-(D.g). By decoupling the effects on BEV features by each DoF of poses, the \textbf{\texttt{DEMA}} solver can individually solve 3 DoF poses by neural matching in a divider-and-conquer manner, reducing the amount of pose hypotheses from $N_x\times N_y \times N_\alpha$ to $N_x+N_y+N_\alpha$. In the experiments, the \textbf{\texttt{DEMA}} solver can achieve a comparable or even better localization accuracy than \textbf{\texttt{FUMA}} solver while significantly reducing around 96\% inference memories required by the pose solver, indicating the proposed \textbf{\texttt{DEMA}} solver an efficient and effective alternative to traditional \textbf{\texttt{FUMA}} solver.

\paragraph{Perception Features v.s. Perception Results}
We analyze the desirable inputs of the \textbf{\texttt{DEMA}} solver by feeding BEV features (Tab. \ref{table:coarse} (D.c), (D.g), (D.h), (D.i), (D.j), and (D.k)) or perception results (Tab. \ref{table:coarse} (D.a) and (D.b)). By comparing the performance of (D.a) and (D.b), we can see that the dimension of perception results (perceived static map mask) is as low as 4, the perception results are not stable without map prior enhancement, so the pose solver using perception results works much worse than the one using BEV features. If we increase the dimension $D_c$ of BEV features fed into the pose solver as (D.h) and (D.j), the localization accuracy consistently improves, indicating that higher-dimensional features are more stable and accurate, but the computational consumption will naturally increase.
When incorporating the \texttt{\textbf{FLORA}} module into BEV perception, the quality of visual perception is significantly enhanced, so the localization performance of pose solvers using both BEV features and perception results is all improved, such as (D.a)-(D.b), (D.c)-(D.g), (D.h)-(D.i), and (D.j)-(D.k). This concludes the necessity of BEV perception enhancement in this work that the E2E localization requires accurate and stable BEV perception as inputs.
By considering both the localization accuracy and computational efficiency, we finally select the \textbf{\texttt{DEMA}} solver (i) using enhanced BEV feature with dimension $D_c=16$ as the default configuration during coarse localization.

\paragraph{Effectiveness of Sampling Step}
We also test the effects of regulating sampling step of both \textbf{\texttt{FUMA}} solver and \textbf{\texttt{DEMA}} solver, as shown in Tab. \ref{table:coarse} (F.b), (F.c), (D.d), and (D.e). Refining sampling step increases the amount of pose hypotheses in both solvers, especially in traditional \textbf{\texttt{FUMA}} solver. The number of pose hypotheses in \textbf{\texttt{FUMA}} solver is increased from $N_x\times N_y \times N_\alpha$ to $8\times N_x\times N_y \times N_\alpha$, resulting in unbearable increase in inference memory. The increased computational burden of \textbf{\texttt{DEMA}} solver is not as significant as \textbf{\texttt{FUMA}} solver, since its pose hypotheses are only increased linearly. Although refining the sampling step improves the localization performance of both \textbf{\texttt{FUMA}} solver and \textbf{\texttt{DEMA}} solver, its improvement is not stable. For example, the estimation accuracy of (D.d) in yaw angle is slightly worse than that of (D.c). Therefore, we select 0.4m / 0.2m / 0.4$\degree$ as the default sampling step in the \textbf{\texttt{DEMA}} solver for both accuracy and efficiency.

\paragraph{Effectiveness of BEV Size} 
Before feeding the BEV features into the \texttt{\textbf{DEMA}} solver, we have compressed the size of BEV features for efficiency. We test the performance of a variant using BEV features with large size (\textit{e.g.}, $H_c\times W_c=400\times 200$), as shown in Tab. \ref{table:coarse} (D.f). It seems that using upsampled BEV features slightly improves the localization accuracy.
We believe that larger features in BEV space can provide more detailed information, which helps the model align the perceived features with the map features.
However, enlarging BEV features takes heavy computational burden for pose estimation. Therefore, we keep using compressed BEV features in the proposed E2E localization network.

\subsubsection{Ablation Analysis about the \textbf{\texttt{POET}} module}
In this paper, we propose to utilize a Transformer-based pose regressor \textbf{\texttt{POET}} to refine the coarse estimation in this fine localization stage. We conduct detailed analysis about its performance.

\begin{table}[!t]
\centering
\caption{The localization performance with adaptive coarse-to-fine localization scheme.}
\renewcommand\arraystretch{1.5}
\begin{threeparttable}
\resizebox{\linewidth}{!}{
\begin{tabular}{l | c | c c c | r}
\toprule
\multirow{2}{*}{Methods} & \multirow{2}{*}{$t_c$} & \multicolumn{3}{c|}{Localization MAE $\downarrow$} & Time${}^1$ \\
& & $x~(cm)$ & $y~(cm)$ & $\alpha~(\degree)$ & (ms) \\
\midrule
Ours (C) & 0 & 14.5 & 12.9 & 0.394 & 6.13 \\
Ours (F) & - & 17.1 & 16.5 & 0.389 & 21.08 \\
Ours (C+F) & 0 & 12.5 & 12.8 & 0.387 & 26.55 \\
\hline
Ours (A) & 0.2 & 14.4 & 12.9 & 0.394 & 6.93  \\
Ours (A) & 0.3 & 14.0 & 12.9 & 0.392 & 11.63 \\
\rowcolor{gray!20}Ours (A) & 0.4 & 13.2 & 12.8 & 0.392 & 19.56 \\
Ours (A) & 0.5 & 12.7 & 12.8 & 0.89 & 24.87  \\
Ours (A) & 0.6 & 12.6 & 12.8 & 0.387 & 26.34 \\
Ours (A) & 0.7 & 12.5 & 12.8 & 0.387 & 26.54 \\
\bottomrule
\end{tabular}
}
\end{threeparttable}
\begin{tablenotes}
    \footnotesize
    \item ${}^1$We measures the consuming time of the localization module.
\end{tablenotes}
\label{table:fine}
\end{table}

\begin{figure*}[!ht]
    \centering
    \includegraphics[width=0.97\linewidth]{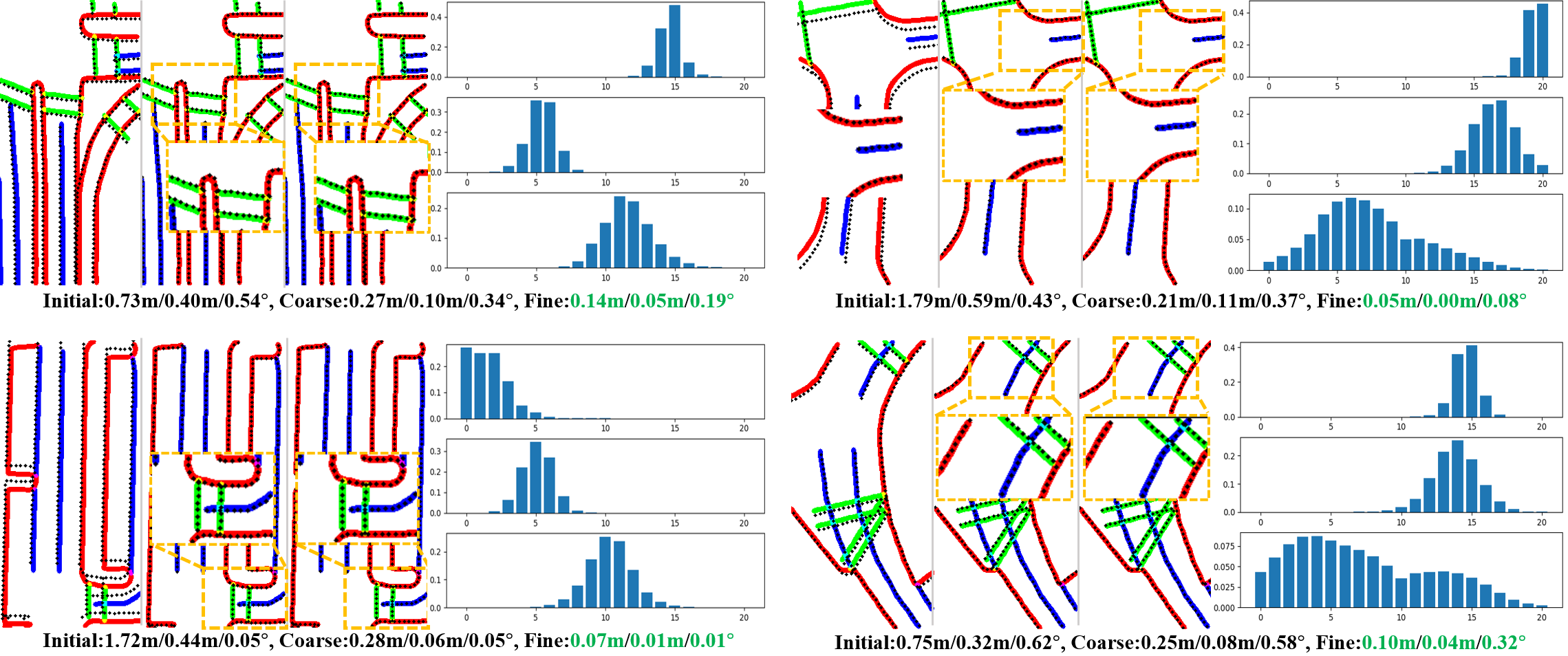}
    \caption{Examples demonstrating the \textbf{\texttt{POET}} module refines the estimation of the \textbf{\texttt{DEMA}} solver. In each case, the map projection using initial pose, coarse pose estimated by \textbf{\texttt{DEMA}} solver, and fine pose estimated by \textbf{\texttt{POET}} module are shown in black circles from left to right; ground-truth HD map masks are drawn in blue, green, and red lines, corresponding to lane dividers, pedestrian crossings, and road boundaries, respectively. Estimated probability distributions from \textbf{\texttt{DEMA}} solver are also provided.}
    \label{fig:fine}
\end{figure*}

\paragraph{Effectiveness of the \textbf{\texttt{POET}} module} 
We first validate the effectiveness of the proposed \textbf{\texttt{POET}} module, as shown in Tab. \ref{table:fine}. In the experiments, we found that the \textbf{\texttt{DEMA}} solver converges rapidly yet fails to improve consistently with extended training,  which stems from the limited resolution of BEV perception and rasterized map encoding.
Using the \texttt{\textbf{POET}} module to refine the coarse estimation from the \textbf{\texttt{DEMA}} solver can effectively further improve the localization accuracy, especially in terms of longitudinal pose estimation, which is necessary for high-level AD systems. We attribute this to the utilized high-precision map features and deep Transformer-based network structures. Although the \textbf{\texttt{POET}} module converges slower and requires more computational resources than the \textbf{\texttt{DEMA}} solver, it has better potential to achieve higher localization accuracy.

Fig. \ref{fig:fine} demonstrates the \textbf{\texttt{POET}} module refines the coarse estimation. Due to the limited resolution of BEV features, the localization accuracy of \textbf{\texttt{DEMA}} solver is near the resolution of BEV perception (0.15m/pixel), which supports spatial alignment between two BEV features. After refinement by the \textbf{\texttt{POET}} module, the localization accuracy of the system is further improved, achieving a centimeter-level high-precision pose estimation.

\paragraph{Effectiveness of the Referenced Coarse Pose Estimation} 
Different from existing Transformer-based pose regressor \cite{bevlocator,du2025rtmap,miao2023poses}, the proposed \textbf{\texttt{POET}} module incorporates the coarse pose estimation from the \textbf{\texttt{DEMA}} solver as a referenced information, as shown in Eqn. \ref{eqn:fine}. To prove the effectiveness of this referenced information, we build a variant (\texttt{Ours(F)}) by removing the coarse pose estimation fed into the \textbf{\texttt{POET}} module, \textit{i.e.}, $\triangle \hat{\textbf{T}}_f=\textbf{\texttt{POET}}(\textbf{F}^e_\mathcal{I},\textbf{F}^V_\mathcal{M})$ and compare the performance, as shown in Tab. \ref{table:fine}.
From the results, we can see that the localization accuracy degrades dramatically without the coarse pose estimation, demonstrating the necessity of the referenced information in the \textbf{\texttt{POET}} module.
In summary, the coarse localization provides referenced information to the fine localization and reduces the difficulty of training Transformer-based network, whereas the fine localization refines the estimation of the coarse localization in the proposed E2E localization network.

\subsubsection{Ablation Analysis about the adaptive coarse-to-fine strategy}
We analyze the effectiveness of the adaptive strategy in the proposed hierarchical localization system.
By regulating the threshold $t_c$, we can control the activation frequency of the \textbf{\texttt{POET}} module, as shown in Fig. \ref{table:fine}. 
As the threshold increases, more and more coarse estimation results are regarded as unreliable, leading to an increase in the frequency of using the POET module to optimize the results. 
Thus, the localization accuracy is consistently improved by the added fine localization stage, but the computational efficiency is sacrificed.
Finally, considering the balance between localization accuracy and efficiency, we select $t_c=0.4$ as the default threshold in the adaptive strategy.

%% file: secs/5-conclusion.tex
\section{Conclusions}

In this paper, we address the visual-to-HD map localization problem by proposing an E2E hierarchical localization network \textbf{\texttt{RAVE}}, which ensures interpretability and computational efficiency while achieving decimeter-level localization accuracy.
Surrounding images and HD map data are converted into BEV features, rasterized and vectorized map features through neural networks, which overcomes the challenges of data association between cross-modal data.
To improve the robustness of vision-based BEV perception, the proposed \textbf{\texttt{FLORA}} module incorporates the misaligned map prior information into the online perceived BEV features, which guarantees stable inputs for subsequent pose estimation.
Then, a hierarchical localization module is proposed to solve vehicle poses, where the \textbf{\texttt{DEMA}} module efficiently estimates coarse pose using rasterized map features, and the \textbf{\texttt{POET}} module refines the results to high precision using vectorized map features.
The proposed network is fully analyzed on the nuScenes dataset \cite{caesar2020nuscenes} and is concluded to be able to achieve high-precision localization with a MAE of 0.132m, 0.128m, and 0.392$\degree$ in longitudinal, lateral positions, and yaw angle. The interpretable \textbf{\texttt{DEMA}} module in coarse localization significantly reduces inference burden, and the \textbf{\texttt{POET}} module ensures the performance.
By regulating the threshold of estimated probability in the adaptive strategy, the proposed network can achieve a good balance among localization accuracy, computational efficiency, and interpretability, which is essential for ego pose estimation for high-level safety-guarantee autonomous driving.